\documentclass[10pt,journal,compsoc]{IEEEtran}
\ifCLASSINFOpdf
  \usepackage[pdftex]{graphicx}
  \graphicspath{{./imgs/}{./jpeg/}}
  \DeclareGraphicsExtensions{.pdf,.jpeg,.png,.bmp,.jpg}
\else
\fi
\usepackage{amsfonts}
\usepackage{amssymb}
\usepackage{color}
\usepackage{bm}
\usepackage{mathrsfs}
\usepackage{multirow}
\usepackage{url}
\usepackage{caption}
\usepackage{color}
\usepackage{multirow}
\usepackage{hyperref}
\usepackage{pifont}
\usepackage{verbatim}  
\newcommand{\scolor}{black}
\newcommand{\ncolor}{black}

\newcommand{\sset}[1]{\bm{\mathcal{\MakeUppercase{#1}}}}
\newcommand{\smat}[1]{\bm{\MakeUppercase{#1}}}			
\newcommand{\sfun}[1]{\mathcal{\MakeUppercase{#1}}}		

\usepackage[cmex10]{amsmath}
\usepackage{algorithm}
\usepackage{algorithmic}
\ifCLASSOPTIONcompsoc
 \usepackage[caption=false,font=normalsize,labelfont=sf,textfont=sf]{subfig}
\else
 \usepackage[caption=false,font=footnotesize]{subfig}
\fi
\usepackage{stfloats}
\begin{document}
%
\title{Deep Affinity Network \\for Multiple Object Tracking}


\author{ShiJie~Sun, Naveed~Akhtar, HuanSheng~Song, Ajmal~Mian, Mubarak~Shah}

\markboth{Journal of \LaTeX\ Class Files,~Vol.~13, No.~9, September~2017}%
{Shell \MakeLowercase{\textit{et al.}}: Bare Demo of IEEEtran.cls for Journals}
%



\IEEEtitleabstractindextext{%

\begin{abstract}
Multiple Object Tracking (MOT) plays an important role in solving many fundamental problems in video analysis {\color{\scolor}and} computer vision. Most MOT methods employ two steps: Object Detection and Data Association. The first step detects objects of interest in every frame of a video, and the second establishes correspondence between the detected objects in different frames to obtain their tracks. Object detection has made tremendous progress in the last few years due to deep learning. However, data association for tracking still relies on hand crafted constraints such as appearance, motion, spatial proximity, grouping etc. to compute affinities between the objects in different frames. 
In this paper, we harness the power of deep learning for  data association in tracking by jointly modeling object appearances and their affinities between different frames in an end-to-end fashion. The proposed Deep Affinity Network~(DAN) learns compact, yet comprehensive features of pre-detected objects at several levels of abstraction, and performs exhaustive pairing permutations of those features in any two frames to infer object affinities. DAN also  accounts for multiple objects appearing and disappearing between video frames.
We exploit the resulting efficient affinity computations to associate objects in the current frame deep into the previous frames for reliable on-line tracking. 
Our technique is evaluated on  popular multiple object tracking challenges MOT15, MOT17 and UA-DETRAC. Comprehensive benchmarking under twelve evaluation metrics demonstrates that our approach is among the best performing techniques on the leader board for these challenges. The open source implementation of our work is available at \href{https://github.com/shijieS/SST.git}{https://github.com/shijieS/SST.git}. 
\end{abstract}


}

\maketitle

\IEEEdisplaynontitleabstractindextext

%
\IEEEpeerreviewmaketitle

\section{Introduction}
\label{sec:Intro}
{\color {\scolor} Tracking multiple objects in videos~\cite{Luo2017} requires detection of objects in individual frames and associating those across multiple frames}.
In {\color {\scolor}Multiple Object Tracking (MOT)}, natural division of the task between object \textit{detection} and  {\color{\scolor} \textit{association}   allows off-the-shelf deep learning based} object detectors~\cite{Ren2017a,Impiombato2015,Redmon2016,Felzenszwalb2008a} to be used for tracking. 
{\color{\scolor} However, the problem of object association is yet to fully benefit from the advances in deep learning due to the peculiar nature of the task of affinity computation for the objects that can be multiple frames apart in video.}

{\color{\scolor}Currently,}
a widely used {\color{\scolor} tracking} {\color{\ncolor}approach}~\cite{Hu2012, Zhang2014, Xiang2015}, \cite{Henschel2018a}, \cite{Kim2015} first computes a representation model of {\color{\scolor} pre-}detected objects in a video. Later, that model is used to estimate {\color{\scolor} affinities} between the  objects across {\color{\scolor} different frames}. 
The approaches following this pipeline exploit several distinct types of representation models, including; appearance models \cite{Kim2015},  \cite{Zhang2016b,Nam2015}, \cite{Bae2014}, motion models~\cite{Breitenstein2009, Fleuret2008}, \cite{Fu2017}, \cite{Kutschbach2017}, and  composite models~\cite{Henschel2018a},~\cite{Baea,Wen2014}. The appearance models focus on computing easy-to-track object features that encode appearances of local regions of objects or their bounding boxes~\cite{Kuo2010, Izadinia2012, Yamaguchi2011}. 
{\color{\scolor} Currently, hand-crafted features are common for appearance modeling. However, such features are not robust to illumination variations and occlusions. Moreover, they also fall short on discriminating distinct objects with relatively high similarity.}   

The motion models encode object dynamics to predict object locations in the future frames. Motion modeling techniques~\cite{Breitenstein2009}, \cite{Shafique2008}, \cite{Yu2007} use linear  models under constant velocity assumption~\cite{Breitenstein2009}. These models exploit the smoothness of object's velocity, position or acceleration in the video frames. Motion model based tracking has also witnessed attempts that account for non-linearity~\cite{yang2012multi} to better represent complexity of {\color{\scolor} the} real-world motions. Nevertheless, both linear and non-linear models fail to handle long inter-frame object occlusions well. Moreover, they are also challenged by the scenarios of complex motions. Hence, composite model based tracking~\cite{Henschel2018a}, \cite{Baea,Wen2014}  aims at striking a balance between motion and appearance modeling. However, such a balance is hard to achieve in  real-world~\cite{Ning2016}. 

\begin{figure*}[t]
\centering
\includegraphics[width=6in]{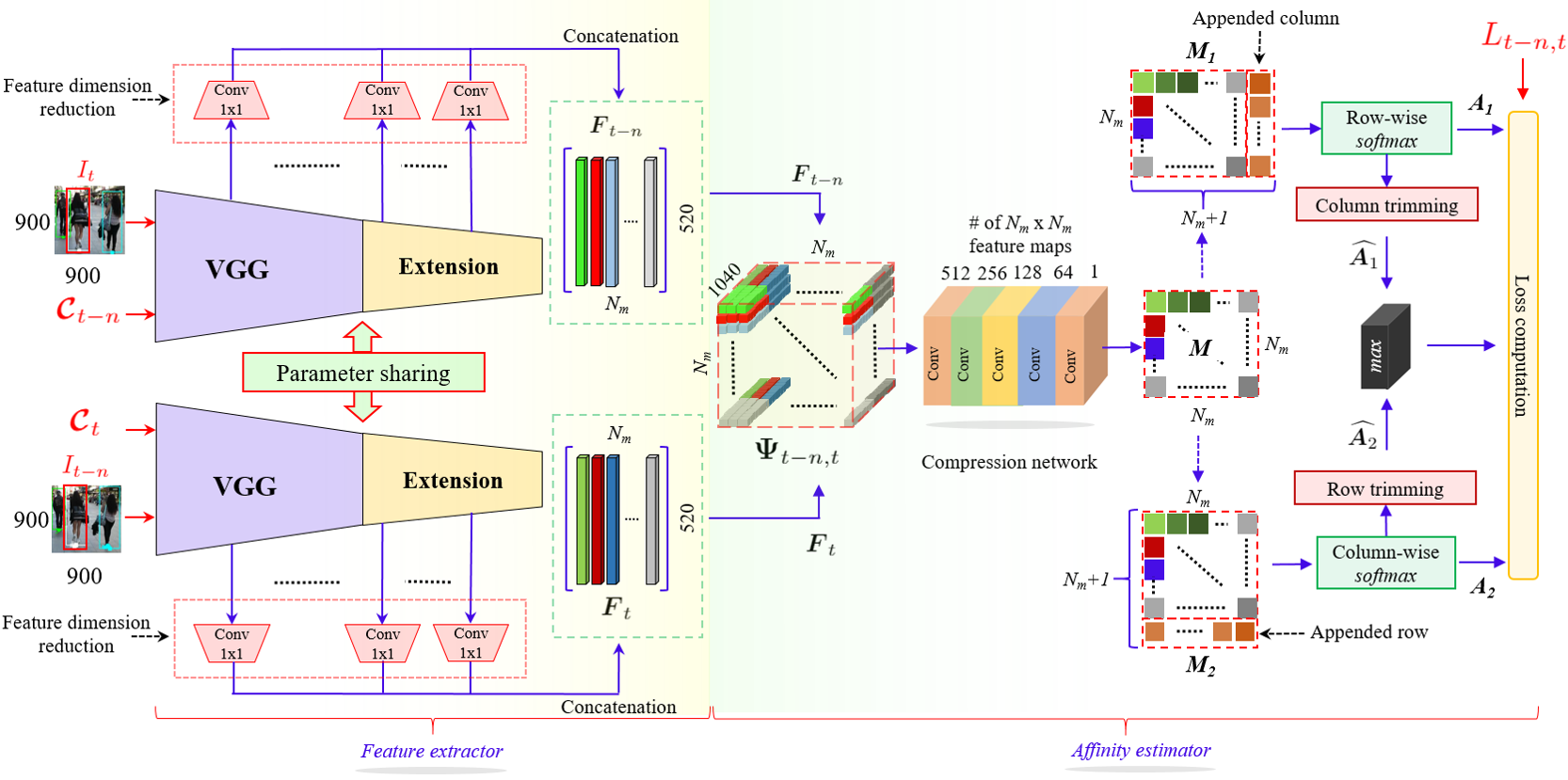}
\caption{Schematics of Deep Affinity Network (DAN): {\color{\ncolor} A pair of video frames $\smat I_t$ and  $\smat I_{t-n}$, which are $n$ time stamps apart} are input to the network along with the sets of centers $\sset C_t$, $\sset C_{t-n}$ of pre-detected objects in those frames. {\color{\scolor} The frame pair is processed by two extended VGG-like   networks with shared parameters}. The {\color{\scolor} number of} feature maps in nine selective layers {\color{\scolor} of these networks} are reduced using $1 \times 1$ convolutional kernels. Compact features extracted from those maps are concatenated to form $520$-dimensional feature vectors. Exhaustive permutations of those vectors in the feature matrices $\smat F_t$ and $\smat F_{t-n}$  are encoded in a tensor  $\boldsymbol \Psi_{t-n,t} \in \mathbb R^{1040\times N_m \times N_m}$, where $N_m$ is the number of objects in each frame.  The tensor $\boldsymbol \Psi_{t-n,t}$ is mapped to a matrix $\smat M \in \mathbb R^{N_m \times N_m}$ using five convolution layers. To account for multiple identities leaving and entering between the frames, $\smat M_1$ and $\smat M_2$ are formed by appending an extra column and an extra row to $\smat M$. Row- and column-wise softmax is performed over $\smat M_1$ and $\smat M_2$ respectively. {\color{\scolor} The resulting matrices $\smat A_1, \smat A_2$ and their column- and row-trimmed variants $\widehat{\smat A_1}, \widehat{\smat A_2}$} are employed in network loss computation using the  ground truth data association matrix $\smat L_{t-n, t}$. {\color{\scolor} Affinity matrix for a pair of frames is predicted using the  matrices $\smat A_1$ and $\smat A_2$ predicted by DAN.}
}
\label{fig:sst}
\vspace{-4mm}
\end{figure*}

To address the challenges of multiple object tracking, we  leverage  the representation power of Deep Learning. We propose a Deep  Affinity Network (DAN), as shown in Fig.~\ref{fig:sst}, that jointly learns target object appearances and their affinities in a pair of {\color{\scolor} video} frames in an end-to-end fashion. 
The appearance modeling accounts for hierarchical feature learning of objects and their surroundings at multiple  levels of abstraction. 
The DAN estimates affinities between the objects in a frame pair under exhaustive permutations of their compact features. The computed affinities also account for the cases of multiple objects entering or leaving the videos between the two frames. 
This is done by enabling the softmax layer of DAN to separately look forward and backward in time for un-identifiable objects in the frame pair.
We propose an appropriate loss function for DAN to account for this unique ability. 
The DAN models object appearance with a two-stream convolutional network with shared parameters, {\color{\scolor} and estimates object affinities in its later layers using the hierarchical features from its initial layers.  } The overall network does not assume the input frame pairs to appear consecutively in a video. This promotes robustness against object occlusions in the induced model.

We exploit the efficient affinity computation by the proposed {\color{\scolor}model} to associate objects in a given video frame to the objects in multiple previous frames for reliable trajectory generation using the Hungarian algorithm~\cite{Munkres1957}.
This results in accurate on-line multiple object tracking.
The proposed approach performs tracking at 6.3 frames per second on the popular MOT15~\cite{Leal-Taixe2015}, MOT17~\cite{MilanL0RS16} and UA-DETRAC~\cite{Wen2015a,Lyu2017} challenge datasets, while surpassing the existing leading approaches on {\color{\ncolor} a} majority of the evaluation metrics. Our technique significantly outperforms its nearest competitors for on-line multiple pedestrian tracking.   

 The rest of the {\color{\scolor}article} is organized as follows. In Section~\ref{sec:rw}, we review the related literature and datasets of multiple object tracking. In Section \ref{sec:m}, we introduce our tracking technique and present the DAN. Section \ref{sec:Exp} details the results of our approach on tracking challenges. {\color{\scolor} We present ablations analysis for the developed technique in Section~\ref{sec:Disc} and draw conclusions in Section \ref{sec:c}.}

\vspace{-2mm}
\section{Related Work} \label{sec:rw}
Multiple Object Tracking (MOT) has attracted significant interest of researchers in recent years. For a general review of {\color{\scolor} this research direction}, we refer to~\cite{Luo2017}, where Luo et al.~organized the existing literature  under multiple criteria and summarized the recent techniques based on  components of {\color{\scolor} the} tracking problem. We also refer to the survey by Emami et al.~\cite{Emami2018} who viewed MOT as an assignment problem and unified a variety of tracking approaches under this formulation.  
Below, we focus more on the contributions that {\color{\scolor} consider} MOT from a data association perspective, and also review recent deep learning based techniques in MOT that relate to our method.


{\color{\scolor}Tracking-by-detection is a generic framework employed by several multiple object trackers~\cite{Tian2018}, {\color{\ncolor}\cite{wen2018learning}, \cite{sheng2018heterogeneous}}. In this framework, the objects are first detected and then associated in different frames.} 
Due to parallel developments of multiple object detectors {\color{\scolor}\cite{Ren2017a}, \cite{Impiombato2015, Redmon2016}, \cite{Felzenszwalb2008a}}, techniques following this line of {\color{\scolor} approach} focus more on the data association aspect of tracking. These techniques can be broadly {\color {\scolor}categorized} as local and global tracking methods.
The local methods~\cite{shafique2005noniterative}, \cite{reid1979algorithm}, \cite{shu2012part}  consider only two frames for data association. This makes them computationally efficient, however, their  performance is  susceptible   to tracking-irrelevant factors such as camera motion, and pose variation etc. 

In contrast to local methods, global techniques \cite{RoshanZamir2012}, \cite{wu2007detection}, \cite{dehghan2015gmmcp} perform data association using a larger number of frames.  {\color{\scolor}Recent methods in this direction} cast data  association into a network flow problem~\cite{Pirsiavash2011}, \cite{butt2013multi}, \cite{Berclaz2011}, \cite{shitrit2014multi}. For instance, Berclaz et al.~\cite{Berclaz2011} solved a constrained flow optimization problem for multiple object tracking, and used {\color{\scolor} the} k-shortest paths algorithm for associating the tracks. Chari et al.~\cite{Chari2015} added a pairwise cost to the min-cost network flow framework and proposed a convex relaxation solution with a rounding heuristic for tracking. Similarly, Shitrit et al.~\cite{shitrit2014multi} used multi-commodity network flow for MOT.
Although popular, this line of methods rely on object detectors  rather  strongly~\cite{Tian2018} that makes {\color{\scolor}them} less desirable for scenarios where occlusions {\color{\scolor} or misdetections} are encountered  frequently. Shu et al.~\cite{shu2012part} handled occlusions up to a scale by extending a part based human detector~\cite{felzenszwalb2010object}.  
There have also been attempts to use dense detections without non-maximum suppression for tracking~\cite{leibe2008coupled}, \cite{tang2015subgraph}. One major goal of these  techniques is to mitigate the problems caused by occlusions and close proximity of the targets that are not handled well by the {\color{\scolor}detectors used for tracking.} 

{\color{\scolor} There are also techniques that mitigate the problems with object detectors in tracking by employing an on-line trainable classifier for the target objects~\cite{Kalal2011}, \cite{hare2011struck},  \cite{wang2011superpixel}}.
Application of on-line classifiers {\color{\scolor} is more beneficial for single object tracking}. However, {\color{\scolor} this notion} has failed to gain popularity in MOT, and has only been successfully applied in limited scenarios~\cite{zhang2013structure}.
There are also instances of MOT under {\color{\scolor} the} tracking-by-detection framework that explicitly focus on mitigating the adverse affects of occlusions. 
{\color{\scolor} For example,} Milan et al.~\cite{Milan2014} employed a continuous energy minimization framework for MOT that incorporates occlusion explicit reasoning and appearance modeling.
To handle occlusions and clutter, Wen et al.~\cite{Wen2014} proposed a data association technique based on undirected hierarchical relation hyper-graph. 
Bochinski et al.~\cite{Bochinski2017a} leveraged the intersection-over-union and predefined thresholds for {\color{\scolor} associating} objects in {\color{\scolor}video} frames. Focusing on computational efficiency, they showed acceptable tracking performance with a speed of up to 100 fps.  
Chen et al.~\cite{Chen2017} proposed a multiple hypothesis tracking method by accounting for scene detections and detection-detection correlations between video frames. {\color{\scolor} Their method can} handle false trajectories while dealing with close object hypotheses.

{\color{\scolor}In the context of deep learning based tracking, models pre-trained for {\color{\ncolor}classification tasks} are popular~\cite{Bertinetto2016, Son2017, Schulter2017, Feichtenhofer2017, Insafutdinov2017}. These models are used to extract object features that are employed for object association in tracking.}
Bertinetto et al.~\cite{Bertinetto2016} proposed to use a fully convolutional Siamese Network~\cite{Chopra2005} for single object tracking.
Similarly, Bea et al.~\cite{Baea} modified the Siamese Network to learn discriminative deep representations for multiple object tracking with object association. They combined on-line transfer learning with the modified network to fine-tune the latter for on-line tracking. 
Son et al.~\cite{Son2017} proposed a quadruplet convolutional neural network that learns to associate objects detected in different video frames. Their network is used by a minimax label propagation method to associate the targets.  
Schulter et al.~\cite{Schulter2017} also proposed a  network that is trained for data association in the context of multiple object tracking.
Feichtenhofer et al.~\cite{Feichtenhofer2017} proposed a multi-function CNN for simultaneous detection and tracking under the detection framework R-FCN~\cite{Dai2016}.
Insafutdinov et al.~\cite{Insafutdinov2017} proposed {\color{\scolor}a} deep learning based method for pose estimation and tracking. They trained a model that groups body parts and tracks a person using the head joint. However, their method {\color{\scolor} underperforms} if heads are  occluded.



{\color{\scolor}Li et al.~\cite{li2018high} proposed a deep learning based technique for single object tracking that uses a Siamese regional proposal network to perform real-time tracking. Similarly, Zhang et al.~\cite{zhang2018learning} proposed a long-term single object tracking framework based on two networks. The first network, that performs regression, generates a series of candidate target objects and their similarity scores for the frames. A second verification network then evaluates these candidates for tracking. Whereas these works contribute specialized deep networks for the tracking problem, their scope is limited to single object tracking.}

{\color{\scolor} In deep learning based tracking, one important aspect of Multiple Object Tracking} research is  collection and annotation of the data itself. {\color{\scolor} This direction has also seen dedicated contributions from the tracking research community}. A summary of the existing multiple object tracking datasets can be found in~\cite{Luo2017}. {\color{\scolor} Among those datasets, two} are particularly relevant to this  work, namely the MOT dataset~\cite{Milan2016} and the UA-DETRAC dataset~\cite{Wen2015a}.     
{\color{\scolor} Introduced by Milan et al.~\cite{Milan2016},} the MOT dataset is used {\color{\scolor} as a benchmark} by a popular recent  multiple object tracking challenge MOT17 (\url{https://motchallenge.net/data/MOT17/}). It  provides a variety of real-world videos for pedestrian tracking. The UA-DETRAC challenge (\url{https://detrac-db.rit.albany.edu/}) uses the large scale UA-DETRAC dataset{\color{\scolor}~\cite{Wen2015a}} that contains videos of traffic in the real-world conditions. We provide further details on both   datasets in Section~\ref{sec:Exp}.
Other popular {\color{\scolor}examples of object tracking datasets} include KITTI~\cite{Geiger2012} and PETS~\cite{Patino2016}.

{\color{\scolor} The representation power of deep learning makes it an attractive choice for appearance modeling based tracking techniques, especially under the tracking-by-detection framework. Nevertheless, the data association component of this framework is yet to fully benefit from deep learning,} 
especially in a manner that {\color{\scolor}  association modeling} is also tailored to  appearance modeling {\color{\scolor} in the overall framework}. This work fills this gap by jointly modeling the appearance of objects in a pair of frames and learning their associations in those frames in an end-to-end manner. The proposed Deep Affinity Network performs comprehensive appearance modeling and uses it further to estimate object affinities in the frame pair. The efficient affinity computation allows our technique to look back {\color{\scolor} into multiple previous frames}  for object association. This keeps the proposed technique robust to object occlusions for tracking.  





\vspace{-2mm}
\section{Proposed Approach} \label{sec:m}


We {\color{\scolor}harness} the representation power of deep learning to perform on-line {\color{\scolor} multiple object} tracking. Central to our {\color{\scolor} technique} is a CNN-based Deep Affinity Network (DAN), see Fig.~\ref{fig:sst}, that jointly models object appearances and their affinities across two different frames that are not necessarily adjacent. 
The overall approach exploits the efficient affinity estimation by  DAN to associate objects in the current frame to those in multiple previous frames to compute  reliable trajectories. We provide a detailed discussion on our approach below.
However, we first introduce notations and conventions for a concise description in the text and figures.

\noindent{\bf Notations:}
\begin{itemize}
\item $\smat I_t$ denotes the $t^{\text{th}}$ video frame under 0-based indexing.

\item $ t-n:t $ denotes an interval from $t-n$ to $t$. 

\item A subscript $t-n, t$ indicates that the entity is computed for the frame pair $\smat I_{t-n}, \smat I_t$.  

\item $\sset{C}_t$ is the set of center locations of the objects in the $t^{\text{th}}$ frame with $\mathcal C_t^i$ as its $i^{\text{th}}$ element.



\item $\smat {F}_t$ is the feature matrix associated with the $t^{\text{th}}$  frame, where its $i^{\text{th}}$ column $F_t^i$ is the feature vector of the $i^{\text{th}}$ object in that video frame.

\item $\boldsymbol\Psi_{t-n,t}$ is a tensor {\color{\scolor} that encodes} all possible pairings of the columns of $\smat {F}_{t-n}$ and $\smat F_{t}$ along its depth dimension.

\item $\smat{L}_{t-n,t}$ denotes a binary data association matrix encoding the correspondence between the objects detected in frames $\smat I_{t-n}$ and $\smat I_t$. If object `1' in $\smat I_{t-n}$ corresponds to the $n^{\text{th}}$ object in $\smat I_t$, then the $n^{\text{th}}$ element of the first row of $\smat{L}_{t-n,t}$ is non-zero. 


\item $\smat{A}_{t-n, t}$ {\color{\scolor} denotes} the affinity matrix {\color{\scolor} that encodes}  similarities between the bounding boxes in $(t-n)^{\text{th}}$ frame and the $t^{\text{th}}$ frame. 


\item $\sset{T}_t$ denotes the set of trajectories or tracks until the $t^{\text{th}}$ time stamp. The $i^{\text{th}}$ element of this set is itself a set of $2$-tuples, containing indices of frames and detected objects. For example, $\mathcal T^i_t = \{(0, 1), (1, 2)\}$ indicates the short track connecting the $1^{\text{st}}$ object in  frame $0$ and the $2^{\text{nd}}$ object in  frame $1$.


\item $\sfun{Z}(\cdot)$ is an operator that computes the number of elements in a set/matrix in its argument. 

\item $\smat\Lambda_{t} \in \mathbb{R}^{\sfun{Z}(\sset{T}_{t-1})\times(\sfun{Z}(\sset{C}_{t})+1)}$ is an accumulator matrix whose coefficient at index $(i,j)$ integrates the affinities of the $i^{\text{th}}$ identity in track set $\sset T_{t-1}$ to the $j^{\text{th}}$ object in the $t^{\text{th}}$ frame over previous $\delta_b$ frames.

\item $N_m$ denotes the allowed maximum number of objects in a frame.
\item $B$ denotes the batch size during training.

\end{itemize}

\noindent {\bf Conventions:}
\
\\
\
The shape of the output at each network layer is described as $Batch \times Channel \times Width \times Height$. For the sake of brevity, we often leave out the $Batch$ dimension. 

\vspace{-2mm}
\subsection{Object detection and localization}
\label{sec:ODL}
The {\color{\scolor} object} detection stage in our approach expects a video frame as {\color{\scolor}} input and outputs a set of bounding boxes for the target objects in that frame. 
{\color{\scolor}For the $t^{\text{th}}$ frame,} we compute object center locations $\sset C_t$ using the {\color{\scolor} output} bounding boxes. 
{\color{\scolor} We evaluate our method (see Section~\ref{sec:Exp})} for different on-line challenges in multiple object tracking that provide their own {\color{\scolor} object detectors}. 
For the MOT17 challenge~\cite{MilanL0RS16}, we use the provided Faster R-CNN~\cite{Ren2017a} and SDP \cite{Yang2016} detectors; for  the MOT15~\cite{Leal-Taixe2015}, we use \cite{Dollar2014};  and for the UA-DETRAC~\cite{Wen2015a,Lyu2017}, we use the EB detector~\cite{Wang2017a}.
Our choices {\color{\scolor} of detectors} are entirely based on the challenges {\color{\scolor} used to evaluate our method}. However, the proposed framework is {\color{\scolor} also} compatible with {\color{\scolor} the} other existing multiple object detectors. 
Since our main contribution is in object tracking and not detection, we refer to the original works for the detectors.

\vspace{-2mm}
\newcommand{\isEq}[1]{\overset{#1}{\sim}}
\subsection{Deep Affinity Network (DAN)}
\label{sec:sst}
We model the appearance of objects in video frames and compute their cross-frame affinities using the Deep Affinity Network (DAN), shown in Fig.~\ref{fig:sst}. To align our discussion with the existing tracking literature, we present the proposed network as two components, {\color{\scolor} namely} (a)~Feature extractor, and (b)~Affinity estimator. {\color{\scolor}However, the overall proposed} network is end-to-end trainable.  
The DAN training requires video frame $\smat I_t$ 
along with its object centers $\sset C_t$; and video frame $\smat I_{t-n}$ 
along with its object centers $\sset C_{t-n}$. We do not restrict the two frames to appear consecutively in video. Instead, we allow them to be $n$ time stamps apart, such that  $n \in \mathbb N \isEq{\text{rand}}[1, N_V] $. Whereas our network is eventually deployed to track objects in consecutive video frames, training it with non-consecutive frames benefits the overall approach in reliably associating objects in a given frame to those in multiple previous frames.  
DAN also requires {\color{\scolor}the} ground truth binary data association matrix 
$\smat L_{t-n, t}$ of the input frame pair 
for computing the network cost during training. 
The inputs to DAN are shown in red color in Fig.~\ref{fig:sst}.
We describe working details of the internal components of our network below after discussing the data preparation.

\subsubsection{Data preparation}
\label{sec:PP}
Multiple object tracking datasets e.g.~\cite{MilanL0RS16}, \cite{Lyu2017} often lack in fully capturing the aspects of camera photometric distortions, background scene variations and other practical factors to which tracking approaches should remain robust. 
For model based approaches, it is important that the training data contains sufficient variations of such tracking-irrelevant factors to induce robustness in the learned models.
Hence, we perform the following preprocessing steps over the available data. 

\begin{enumerate}
\item \emph{Photometric distortions}: Each pixel of a video frame is first scaled by a random value in the range $[0.7, 1.5]$. The resulting image is converted to $HSV$ format, and its Saturation channel is again scaled by a random {\color{\scolor}value} in $[0.7, 1.5]$. The frame is then converted back to $RGB$ format and rescaled by a random {\color{\scolor}value} in the same range. This process of photometric distortion is similar to~\cite{Liu2016}, that also inspires the used range values. 
\item \emph{Frame expansion}: We expand the frames by a random ratio sampled in the range $[1, 1.2]$. The expansion results in increasing {\color\scolor the} frame sizes. {\color{\scolor} To achieve the new size, we pad the original frame with extra pixels. The value of these extra pixels is set to the mean pixel value of the training data.}
\item \emph{Cropping}: We  crop the frames using cropping ratios randomly sampled in the range $[0.8, 1]$. We keep only those crops that contain the center points of all the detected boxes in the original frames. 
\end{enumerate}

Each of the above steps is applied to the frame pairs sequentially with a probability 0.3. The frames are then resized to fixed dimensions $H \times W \times 3$, and horizontally fliped with a probability of 0.5.
The overall strategy of modifying the training data is inspired by Liu et al.~\cite{Liu2016} who alter images to train an object detector. 
However, different from~\cite{Liu2016}, we simultaneously process two frames by applying the above-mentioned transformations to them. 

\begin{figure}
\centering
\includegraphics[width=3in]{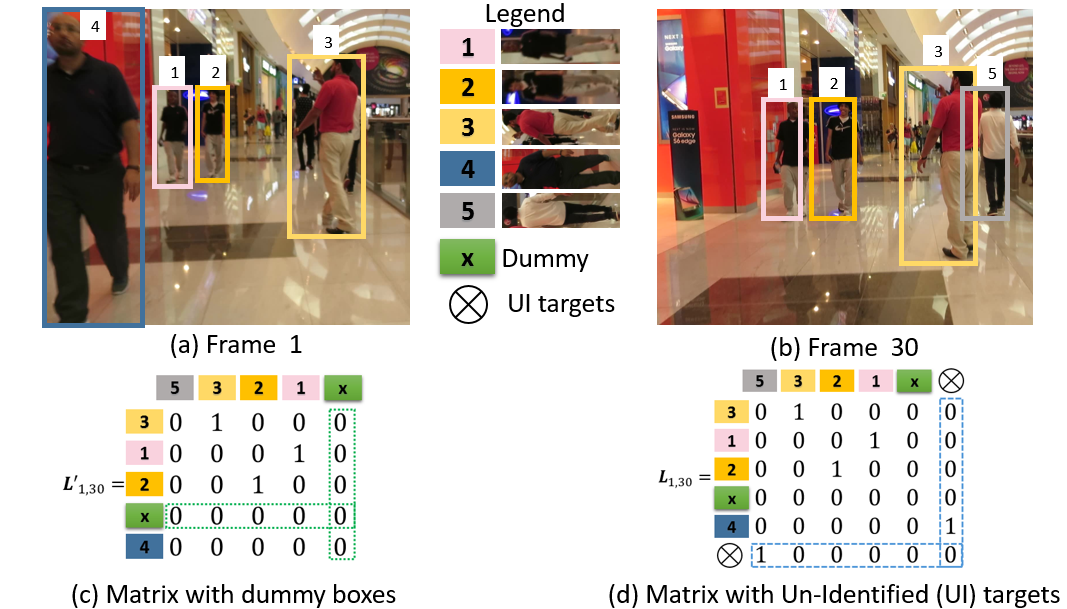}
\caption{Illustration of data association matrix between frame 1 and 30, with $N_m =5$. Frame 1 and 30 in (a) and (b) jointly contain 5 detected objects. (c) Creation of an intermediate matrix that considers \textit{dummy} objects (rows and columns with zeros) to achieve $N_m = 5$ per frame. (d) Augmentation with extra column and row to include Un-Identified (UI) targets (objects leaving and entering respectively) between the two frames.}
\label{fig:label}
\end{figure}

The resulting processed frames are used as inputs to DAN along with the associated object centers computed by {\color{\scolor} the} detector. 
{\color{\scolor} This stage also accounts for object occlusions. Fully occluded objects in the training data are ignored by our training procedure at this point.}
{\color{\ncolor} We set the visibility threshold of 0.3 to treat an object as fully occluded. The remaining partially occluded objects become the positive samples of objects with occlusions.}
We compute data association matrices for the frames by introducing an upper bound $N_m$ on the maximum number of objects allowed in a given frame. In our experiments, $N_m = 80$ proved a generous bound for the benchmark challenges. 
For consistency, we introduce additional rows and columns (with all zeros) in the data association matrix corresponding to \emph{dummy} bounding boxes in each video frame so that all the frames eventually contain $N_m$ objects and the matrix size is $N_m \times N_m$.  
Figure~\ref{fig:label} illustrates {\color{\scolor}our approach for} the construction of data association matrix, where we let $N_m = 5$ for simplicity. {\color{\scolor} In the figure,} frames 1 and 30 contain four detected persons each, amounting to five distinct identities. {\color{\scolor} Figure}~\ref{fig:label}c shows the construction of an intermediate matrix $\smat{L}'_{1,30}$ for the frames with {\color{\scolor}a} row and {\color{\scolor}a}  column for dummy bounding boxes. In  Fig.~\ref{fig:label}d, the association matrix is  augmented with an extra row and {\color{\scolor}an extra} column, labeled as \emph{un-identified} targets (UI targets). The augmented column accounts for currently tracked objects leaving the video and the augmented row accounts for {\color{\scolor}the}  new objects entering the video. 
We eventually use the form of ground truth association matrix shown in Fig.~\ref{fig:label}d to train the DAN.
In the shown illustration, the last column of the augmented matrix has a 1 for object-4 because it has left and the last row has a 1 for object-5 because it has appeared in Frame 30. Notice that, using this convention, DAN is able to account for multiple objects leaving and entering the video i.e. by placing 1's at multiple rows in the last column and by placing 1's at multiple columns in the last row, respectively. 
After data preparation, the entities available as inputs to our network are summarized in Table~\ref{tab:input_item}. 

\begin{table}[t]
\centering
\caption{The input data for training DAN. The tensor dimensions are expressed as Channels $\times$ Height $\times$ Width.}
\label{tab:input_item}
\begin{tabular}{ll}
\hline
Input         	&        Dimensions/Size\\
\hline
$\smat I_{t-n}$, $\smat I_{t}$        	&        $ 3 \times H \times W$ \\
$\smat{L}_{t-n, t}$ &		 $ 1 \times (N_m+1) \times (N_m+1)$ \\
$\sset{C}_{t}$, $\sset{C}_{t-n}$    	&        $N_m $ \\
\hline
\end{tabular}
\vspace{-3mm}
\end{table}

\vspace{-2mm}
\subsubsection{Feature extractor} 
\label{sec:fes}
We refer to the first major component of DAN as \textit{feature extractor} {\color{\scolor}given} its functionality.
This sub-network models comprehensive, yet compact features of the detected objects in video frames. As shown in Fig.~\ref{fig:sst}, the feature extraction is performed by passing pairs of video frames and object centers through two streams of convolution layers. These streams share the model  parameters  in our implementation, whereas their architecture is inspired by VGG16 network~\cite{Simonyan2015}. We use the VGG architecture after converting its fully-connected and softmax layers to convolution layers. This modification is made because the spatial features of objects, which are of more interest in our task, are better encoded by convolution layers. Compared to the original VGG16, the input frame size for our network is much larger (i.e. $3\times 900\times 900 $) due to the nature of the task at hand and the available tacking datasets. Consequently, we are still able to compute $56 \times 56$ feature maps after the last layer of the modified VGG network. In Fig.~\ref{fig:sst}, we index the last layer of VGG as the $36^{\text{th}}$ layer under the convention that Batch Normalization~\cite{ioffe2015batch} and ReLU activations~\cite{Nair2010} are counted as separate layers. 
We refer to \cite{Simonyan2015} for the details on the original VGG architecture.

We reduce the spatial dimensions of our feature maps beyond  $56 \times 56$, after the VGG layers, by introducing further convolution layers. The 36-layer \textit{Extension} network gradually reduces feature maps to size $3 \times 3$. {\color{\scolor}Our choice of gradually reducing the feature maps to size $3 \times 3$ is empirical. We conjecture that it results in better performance because it ensures comprehensive appearance modeling at multiple levels of abstraction. Due to their large receptive fields, the latter layers of the Extension sub-network are able to better model object surroundings, which helps in the overall performance of our technique.} The architectural details of the extension network  are provided in the top half of Table~\ref{tab:extra_net}. The table counts the output of VGG network as input to the first layer of the extension that is indexed 0. Since the Batch Normalization and ReLU are counted as separate layers, the first column of the table increments indexes with step size~3.  

\begin{table}[t]
  \centering
  \tabcolsep=3.5pt
  \caption{Architectural details of the \emph{Extension} \& \emph{Compression} networks in Fig.~\ref{fig:sst}:  I.C denotes the number of input channels for a layer, O.C is the number of output channels, S is the stride size, B.N (Y/N) indicates if the Batch Normalization is applied; and ReLU (Y/N) indicates if the ReLU activation is used. Strides and Paddings are the same in both spatial dimensions.}
  \label{tab:extra_net}
  \tabcolsep=0.07cm
    \begin{tabular}{ccccccccc}
    \hline
    \textbf{Sub-network}	&\textbf{Index} & \textbf{I.C} & \textbf{O.C} & \textbf{Kernel} & \textbf{S} & \textbf{Padding} & \textbf{B.N} & \textbf{ReLU} \\
    \hline
    \multirow{5}{*}{}
    &	0     & 1024  & 256   & $1 \times 1$ & 1 & 0  & Y & Y\\
    &	3     & 256   & 512   & $3 \times 3$ & 2 & 1  & Y & Y\\
    &	6     & 512   & 128   & $1 \times 1$ & 1 & 0  & Y & Y\\
    &	9     & 128   & 256   & $3 \times 3$ & 2 & 1  & Y & Y\\
    \multirow{1}{*}{Extension}
    &	12    & 256   & 128   & $1 \times 1$ & 1 & 0  & Y & Y\\
    \multirow{1}{*}{(Feature extractor)}
    &	15    & 128   & 256   & $3 \times 3$ & 2 & 1  & Y & Y\\
    &	18    & 256   & 128   & $1 \times 1$ & 1 & 0  & Y & Y\\
    &	21    & 128   & 256   & $3 \times 3$ & 2 & 1  & Y & Y\\
    &	24    & 256   & 128   & $1 \times 1$ & 1 & 0  & Y & Y\\
    &	27    & 128   & 256   & $3 \times 3$ & 2 & 1  & Y & Y\\
    &	30    & 256   & 128   & $1 \times 1$ & 1 & 0  & Y & Y\\
    &	33    & 128   & 256   & $3 \times 3$ & 2 & 1  & Y & Y\\
    \hline\hline
    \multirow{2}{*}{}
    &	0     & 1040  & 512  	& $1 \times 1$ 	& 1 & 0 &	Y	&	Y\\
    &	3     & 512   & 256   	& $1 \times 1$	& 1 & 0 &	Y	&	Y\\
    \multirow{1}{*}{Compression}
    &	6     & 256   & 128   	& $1 \times 1$ 	& 1 & 0 &	Y	&	Y\\
    \multirow{1}{*}{(Affinity estimator)}
    &	9     & 128   & 64   	& $1 \times 1$ 	& 1 & 0 &	N	&	Y\\
    &	11    & 64    & 1    	& $1 \times 1$ 	& 1 & 0 &	N	&	Y\\
    \hline
    \end{tabular}%
    \vspace{-3mm}
\end{table}%

Knowing the object center locations in the input frames (as $\sset C_t$ and $\sset C_{t-n}$) allows us to extract center pixels of the objects as their representative features. We extend this notion to the feature maps of our network and sample the maps of different convolution layers at object centers after accounting for reduction in {\color{\scolor}their} spatial dimensions. Due to multiple sequential convolution operations along the network layers, the sampled vectors represent object features at different levels of abstraction. 
These are highly desirable characteristics of features for tracking.
To ensure such features are sufficiently expressive, it remains imperative to learn a large number of feature maps. However, this would make the comprehensive feature vector formed by combining features from multiple layers too large to be practically useful.



We side-step this issue by reducing the number of feature maps of nine empirically selected layers in our network. 
This reduction is performed with additional convolution layers branching out from the two main streams of the feature extractor (see Fig.~\ref{fig:sst}). 
The additional layers use $1 \times 1$ convolution kernels for dimensionality reduction.
Table~\ref{tab:selected_layers} lists indices of the selected nine layers along the number of channels at  input and output of the convolution layers performing dimensionality reduction. 
Our network concatenates feature vectors from the selected nine  layers  to form a $520$-dimensional vector for a detected object. By allowing $N_m$ detections in the $t^{\text{th}}$ frame, we obtain a feature matrix $\smat F_t \in \mathbb R^{520 \times N_m}$. Correspondingly, we also construct $\smat F_{t - n} \in \mathbb R^{520 \times N_m}$ for the $(t- n)^{\text{th}}$ frame.
Recall that these matrices also contain features for dummy objects that actually do not exist in the video frames. We implement these features as zero vectors. 

\begin{table}[t]
  \centering
  \caption{Details of \emph{feature dimension reduction} layers in Fig.~\ref{fig:sst}:  3 layers are selected from VGG network. 6 layers are selected from the Extension network.  I.C~denotes the number of input channels. S.D~indicates spatial dimensions of feature maps. O.C~is the number of output channels. }
  \label{tab:selected_layers}
    \begin{tabular}{lccccccc}
    \hline
    \multicolumn{1}{l}{\textbf{Sub-network}} & \textbf{ Layer index} & \textbf{I.C} & \textbf{S.D} & \textbf{O.C} \\ \hline
    \multirow{3}[0]{*}{VGG} & 16    & 256   & 255x255 &	60\\
          & 23    & 512   & 113x113 	&	80	\\
          & 36    & 1024  & 56x56   	&	100	\\
    \hline
    \multirow{6}[0]{*}{Extension network} 
          & 5     & 512   &  28x28		&	80\\
          & 11    & 256   &  14x14		&	60\\
          & 17    & 256   &  12x12		&	50\\
          & 23	  & 256   &  10x10		&	40\\
          & 29    & 256  &  5x5  		&	30\\
          & 35    & 256   &  3x3  		&	20\\
    \hline
    \end{tabular}%
    \vspace{-3mm}
\end{table}%

\subsubsection{Affinity estimator}
\label{sec:matcher}
The objective of this component of DAN is to encode affinities between the objects using their extracted features. 
To that end, the network arranges the columns of $\smat F_t$ and $\smat F_{t-n}$ in a tensor  $\boldsymbol\Psi \in \mathbb R^{N_m \times N_m \times (520 \times 2)}$, such that the columns of the two feature matrices are concatenated along the depth dimension of the tensor in $N_m \times N_m$ possible permutations, see Fig.~\ref{fig:sst} for illustration.  We map this tensor onto a matrix $\smat M \in \mathbb R^{N_m \times N_m}$  through a \emph{compression network} that uses 5 convolution layers with $1 \times 1$ kernels. Specifications of this network are given in the bottom half of Table~\ref{tab:extra_net}.

The architecture of the compression network is inspired by the physical significance of its input and output signals. The network maps a tensor {\color{\scolor} that encodes} combinations of object features to a matrix that codes similarities between the features (hence, the objects).
Thus, it performs a gradual dimension reduction along the depth of the input tensor with convolutional kernels that do not allow neighboring elements of feature maps to influence each other.
For a moment, consider a forward-pass through the DAN until computation of {\color{\scolor} the matrix} $\smat M \in \mathbb R^{N_m \times N_m}$. {\color{\scolor} The remainder of our  network must compare this matrix to the ground truth data association matrix $\smat L_{t-n, t} \in \mathbb R^{(N_m +1) \times (N_m+1)}$ for loss computation}. 
However, unlike $\smat L_{t-n, t}$, $\smat M$ does not account for the objects {\color{\scolor} that enter or leave the video between the two  input frames}.
To take care of those {\color{\scolor} objects}, we also append an extra column and an extra row to $\smat M$ to form matrices $\smat M_1 \in \mathbb R^{N_m \times (N_m + 1)}$  and $\smat M_2 \in \mathbb R^{ (N_m + 1) \times N_m}$, respectively. This is analogous to the augmentation of $\smat L_{t-n, t}$ explained in Section~\ref{sec:PP}. However, here we separately append the row and column vectors to $\smat M$ in order to keep  the loss computation well-defined and physically interpretable.
The vectors appended to $\smat M$ take the form ${\bf v} \in \mathbb R^{N_m} = \gamma {\bf 1}$, where ${\bf 1}$ is a vector of ones, and $\gamma$ is a hyper-parameter of the proposed DAN. 

\vspace{2mm}
\noindent{\bf Network loss:}

{\color{\scolor} In our formulation}, the $m^{\text{th}}$ row of $\smat M_{1}$ associates the $m^{\text{th}}$ identity in frame $\smat I_{t-n}$ to $N_m+1$ identities in frame $\smat I_t$,  where $+1$ results {\color{\scolor}from the unidentified (UI) objects in $\smat I_t$.} 
We fit a separate probability distribution over each row of $\smat M_1$ by applying a \textit{row}-wise softmax operation over the matrix. Thus,  {\color{\scolor} a row} of the resulting matrix $\smat A_{1} \in \mathbb R^{N_m \times (N_m +1 )}$ encodes probabilistic associations between {\color{\scolor}an} object in frame $\smat I_{t-n}$ and all identities in frame $\smat I_t$.  
We correspondingly apply a \textit{column}-wise softmax operation over $\smat M_2$ to compute $\smat A_2 \in \mathbb R^{(N_m +1) \times N_m}$,  whose columns signify similar  \textit{backward} associations from frame $\smat I_t$ to $\smat I_{t-n}$. {\color{\scolor} It is emphasized that our probabilistic object association}  allows for multiple objects entering or leaving the video between {\color{\scolor} the two frames of interest}. 
The maximum number of {\color{\scolor}objects that can  enter or leave video} is upper-bounded by $N_m$ - the maximum number of allowed objects in a frame.


We define the loss function for DAN with the help of four sub-losses, referred to as 1) {\color{\ncolor}Forward-direction loss} $\mathcal L_{f}:$ that encourages correct identity association from  $\smat I_{t-n}$ to  $\smat I_t$. 2)~{\color{\ncolor}Backward-direction loss} $\mathcal L_{b}:$ that ensures correct associations from $\smat I_t$ to  $\smat I_{t-n}$. 3) Consistency loss $\mathcal L_c:$ to rebuff any  inconsistency between $\mathcal L_{f}$ and $\mathcal L_{b}$. 4) Assemble loss $\mathcal L_{a}:$ that suppresses non-maximum forward/backward associations for affinity predictions. 
Concrete definitions of these losses are provided below: 
\begin{align}
\mathcal L_f(\smat{L}_1, {\smat A_1}) &= \frac{\sum \limits_{\text{{\color{\scolor}coeff}}}\left( \smat{L}_1 \odot \left(-\log {\smat A_1} \right) \right)}{\sum \limits_{{\color{\scolor}\text{coeff}}} (\smat L_1)}, \\
\mathcal L_b(\smat{L}_2, {\smat A_2}) &= \frac{\sum \limits_{{\color{\scolor}\text{coeff}}}\left( \smat{L}_2 \odot \left(-\log {\smat A_2} \right) \right)}{\sum \limits_{{\color{\scolor}\text{coeff}}} (\smat L_2)}, \\
\mathcal L_c (\widehat{\smat A_1}, \widehat{\smat A_2}) &= || \widehat{\smat A_1} - \widehat{\smat A_2} ||_1, \\
\mathcal L_a(\smat{L}_3, \widehat{\smat A_1}, \widehat{\smat A_2}) &= \frac{\sum \limits_{{\color{\scolor}\text{coeff}}}\left( \smat{L}_3 \odot \left(-\log (\max(\widehat{\smat A_1}, \widehat{\smat A_2})) \right) \right)}{\sum \limits_{{\color{\scolor}\text{coeff}}} (\smat L_3)}, \\
\mathcal L &= \frac{\mathcal L_f + \mathcal L_b + \mathcal L_a + \mathcal L_c}{4}.
\end{align}
In the above equations, $\smat L_1$ and $\smat L_2$ are the trimmed versions of $\smat L_{t-n,t}$ constructed by ignoring the last row and {\color{\scolor} the last} column of {\color{\scolor}$\smat L_{t-n,t}$}, respectively. $\widehat{\smat A_1}$ and $\widehat{\smat A_2}$ denote the matrices $\smat A_1$ and $\smat A_2$  trimmed to the size $N_m \times N_m$ by respectively dropping the last column and the last row. Similarly, $\smat L_3$ drops out both {\color{\scolor} the} last row and {\color{\scolor}the last} column of $\smat L_{t-n,t}$. The operator $\odot$ denotes the Hadamard  product, {\color{\scolor} and $\sum \limits_{\text{coeff}}(.)$ sums up all coefficients of the matrix in its argument to a scalar value}. The \textit{max} and \textit{log}  operations are also performed element-wise.
{\color{\scolor} In the above defined losses, one noteworthy observation is regarding separate losses for `Forward' and `Backward' associations. We define these losses using two individual matrices $\boldsymbol{A}_1$ and $\boldsymbol{A}_2$. With an extra column and `row'-wise softmax operation applied to its coefficients, $\boldsymbol{A}_1$ is inherently dissimilar to $\boldsymbol{A}_2$ that is computed with a `column'-wise softmax operation and has  an extra row. To account for different matrix sizes and more importantly, the difference in the physical significance of their coefficients, we define individual losses for them.}

We compute the final loss $\mathcal L$ as the mean value of the four sub-losses. The overall cost function of our network is defined as the Expected value of the training data loss. 
The afore-mentioned four sub-losses are carefully designed for our problem. 
In the Forward and Backward {\color{\ncolor}direction} losses, instead of forcing $\smat A_q$, where $q \in \{1,2\}$; to approximate corresponding $\smat L_q$ by using a distance metric, we maximize the probabilities encoded by the relevant coefficients of $\smat A_q$. We argue that this strategy is more sensible than minimizing a distance between a binary matrix $\smat L_q$ and a probability matrix $\smat A_q$. 
Similarly, given the difference between $\widehat{\smat A_1}$ and $\widehat{\smat A_2}$ is expected to be small, we employ $\ell_1$-distance instead of {\color{\scolor}the} more commonly used $\ell_2$-distance for the Consistency loss.   
Once the DAN is trained, we use it to compute the affinity matrix for an input frame pair as $\smat A \in \mathbb R^{N_m \times N_m+1} = \mathcal A(\max(\widehat{\smat A_1}, \widehat{\smat A_2}))$, where $\mathcal A(.)$ appends the $(N_m + 1)^{\text{th}}$ column of $\smat A_1$ to the matrix in its argument. The max operation used in our definition of the affinity matrix $\smat A$ also justifies the maximization performed to compute the Assemble loss. Thus, the four sub-losses defined above are complementary that result in a systematic approximation of the ground truth data association. 

\vspace{-2mm}
\subsection{DAN deployment}
\label{sec:dep}
\begin{figure*}[t]
\centering
\includegraphics[width=4.5in]{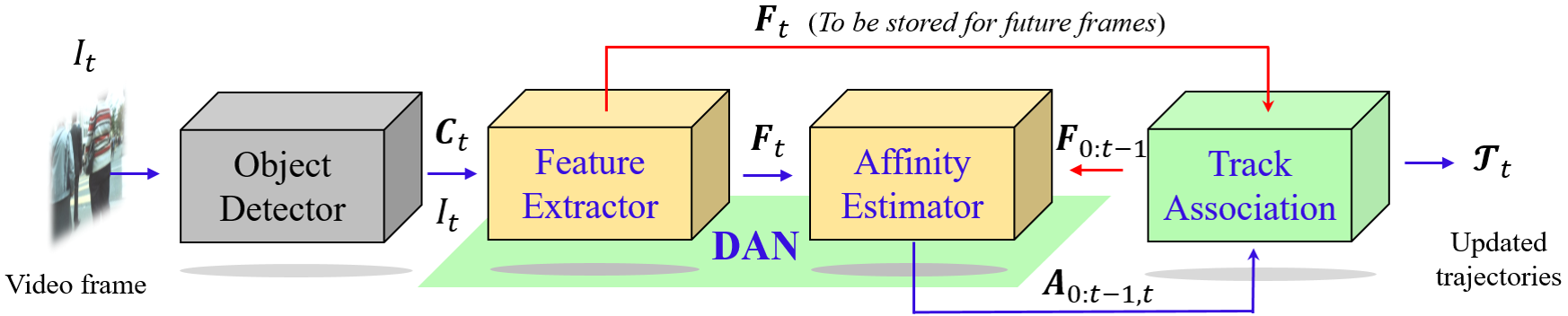}
\caption{Deep tracking with DAN deployment: For the $t^{\text{th}}$ frame $\smat I_t$, the object centers $\mathcal C_t$ provided by the detectors are used to compute the feature matrix $\smat F_t$ with \textit{one} stream Feature Extractor of DAN. The $\smat F_t$ is paired with each of the last $t$ feature matrices $\smat F_{0:t-1}$, and each pair is processed by the Affinity Estimator to compute the same number of affinity matrices $\smat A_{0:t-1,t}$. The $\smat F_t$ is also stored for computing affinity matrices in the future.  The trajectory set $\mathcal T_t$, is updated by associating the current frame with $t$ previous frames using the computed affinity matrices.   
}
\label{fig:framework}
\vspace{-2mm}
\end{figure*}

Whereas the \textit{feature extractor} component of DAN is trained as a two-stream network, it is deployed as a one-stream model in our approach. This is possible because the parameters are shared between the two streams. In Fig.~\ref{fig:framework}, we illustrate the deployment of DAN by showing its two major components separately. The network expects a single frame $\smat I_t$ as its input, along the object center locations $\mathcal C_t$. The feature extractor computes the feature matrix $\smat F_t$ for the frame and passes it to the \textit{affinity estimator}. The latter uses the feature matrix of a previous frame, say $\smat I_{t-n}$ to compute the permutation tensor $\boldsymbol\Psi_{t-n,t}$ for the frame pair. The tensor is then mapped to an affinity matrix by a simple forward pass through the network and a concatenation operation, as described above. Thus each frame is passed through the object detector and feature extractor only once, but the features are used multiple times for computing affinities with multiple other frames in pairs.


\vspace{-2mm}
\subsection{Deep track association}
\label{sec:rec}
To associate objects in the current frame with multiple previous frames, we store feature matrices of the frames with their time stamps.
After frame 0, that initializes our approach and results in $\smat F_0$, we can compute affinities between the objects in the current frame and those in any previous frame using their feature matrices. 
As illustrated in Fig.~\ref{fig:framework}, the computed affinity matrices are used to update trajectory sets by looking back deep into the previous frames.

Deep track association  is performed as follows.
We initialize our track set $\sset T_0$ with as many trajectories as the number of detected objects in $\smat I_0$. A trajectory here is a set of 2-tuples, each containing the time stamp of the frame and the object identity.
The trajectory set is updated at the $t^{\text{th}}$ time stamp with the help of Hungarian algorithm~\cite{Munkres1957} applied to an accumulator matrix $\smat\Lambda_{t} \in \mathbb{R}^{\sfun{Z}(\sset{T}_{t-1})\times(\sfun{Z}(\sset{C}_{t})+1)}$.
{\color{\scolor} The Hungarian algorithm solves an assignment problem using combinatorial optimization. 
We formulate this problem using the accumulator matrix, whose rows represent the existing object identities and columns encode their similarities to the objects in the current frame. The Hungarian algorithm computes unique assignments of the objects to the identities. This is done by maximizing the affinities between the current frame objects and the objects already assigned to the identities in previous frames.}
The used accumulator matrix integrates the affinities between  objects in the current frame and the previous frames. 
A coefficient of $\smat\Lambda_{t}$ at index $(i,j)$ is the sum of affinities of the $i^{\text{th}}$ identity in the track set $\sset T_{t-1}$ to the $j^{\text{th}}$ object in the $t^{\text{th}}$ frame for the previous $\delta_b$ frames, where $\delta_b$ is a parameter of our approach.  



At each time stamp, we are able to efficiently compute up to $\delta_b$ affinity matrices using the DAN to look back into the existing tracks. We let $\delta_b = t$ in Fig.~\ref{fig:framework} for {\color{\scolor}simplicity}.
One subtle issue in successfully applying the Hungarian algorithm to our problem is that we allow multiple objects to leave a video between its frames. Therefore, multiple trajectories could be assigned to the \textit{single} {\color{\scolor} Un-Identified} target column in our accumulator matrix (inherited from the affinity matrices).
We handle this issue by repeating the last column of $\boldsymbol \Lambda_t$ until every trajectory in $\sset T_{t-1}$ gets assigned to a unique column of the augmented matrix\footnote{The number of columns of \textit{augmented} $\boldsymbol \Lambda_t$ are allowed to exceed $N_m + 1$. However, it does not have any ramifications as that matrix is only utilized by the Hungarian algorithm.}. This ensures that all un-identified trajectories can be mapped to un-identified objects.

Overall, our tracker is an on-line approach in the sense that it does not use future frames to predict object trajectories. Hence, it can be used with continuous video streams. One practical issue in such cases is that very long tacks may result over large time intervals. The parameter $\delta_b$ also bounds the maximum number of time stamps we associate with a track. We remove the oldest node in a track if the number of frames for its trajectory exceed{\color{\scolor}s} $\delta_b$.
Similarly, for the objects disappearing from a video, we allow a waiting time of  $\delta_w$ frames before removing its trajectory from the current track set.
We introduce these parameters in our framework for purely pragmatic reasons. Their values  can be adjusted according to the on-board computational and memory capacity of a tracker.

\vspace{-2mm}
\section{Experiments}\label{sec:Exp}
In this section, we evaluate the proposed approach on three well-known multiple object tracking challenges, namely Multiple Object Tracking 17 (MOT17)~\cite{Milan2016}, Multiple Object Tracking 15 (MOT15)~\cite{Leal-Taixe2015} and UA-DETRAC~\cite{Wen2015a,Lyu2017}. All these are on-line challenges where tracking results are computed by a hosting server once a new technique  is submitted for evaluation. Annotated training data is provided for learning the models, however, labels of the test data remain {\color{\scolor}undisclosed}. The servers perform comprehensive evaluation of the submitted techniques using several standard metrics. {\color{\scolor} We first report the implementation details of the proposed technique, followed by the on-line challenges;  their training datasets and the used evaluation metrics.} The performance of our approach and its comparison to other techniques currently on the leader board is also presented.

\vspace{-2mm}
{\color{\scolor}\subsection{Implementation details}}
\label{sec:Imp}
We implement the Deep Affinity Network (DAN) using the Pytorch framework~\cite{Paszke2017}. Training is performed on NVIDIA GeForce GTX Titan GPU. Hyper-parameters of DAN are  optimized with the help of MOT17 dataset, for which we specify a validation set to train our model. MOT17 is selected for parameter optimization due to its manageable size. 
The hyper-parameter values finally used in our implementation are as follows. Batch size $B = 8$, number of training epochs per model =  120,  number of maximum objects allowed per frame $N_m = 80$, and $ \gamma = 10$. We let $N_V = 30$. Our network {\color{\scolor}has an} input frame size of $900 \times 900$. All the training and testing data is first resized to these dimensions before passing it through the network. We use the SGD optimizer~\cite{zhang2015deep} for training the DAN, for which we respectively use 0.9 and 5e-4 for the momentum and weight decay parameters. We start the learning process with 0.01 as the learning rate, which {\color{\scolor}is}  decreased to $1/10^{\text{th}}$ of the previous value at epochs 50, 80 and 100.
Once the network is trained, we still need to decide on the values of the parameters $\delta_b$ and $\delta_w$. 
We select the best values of these parameters with a grid search for optimal MOTA metric on our validation set. We use multiples of three in the range [3,30] to form the grid
, based on which $\delta_w = 12$ and $\delta_b = 15$ are selected in the final implementation.

\vspace{-1mm}
\subsection{Multiple Object Tracking 17 (MOT17)}
\label{sec:MOT17}

The Multiple Object Tracking 17 (MOT17)~\cite{Milan2016} is among the latest on-line challenges in tracking. Similar to its previous version MOT16~\cite{MilanL0RS16}, this challenge contains seven different indoor and outdoor scenes of public places with pedestrians as the objects of interest.
A video for each scene is divided into two clips, one for training and the other for testing. The dataset provides detections of objects in the video frames with three detectors, namely SDP~\cite{Yang2016}, Faster-RCNN~\cite{Ren2017a} and DPM~\cite{felzenszwalb2010object}. 
The challenge accepts both on-line and off-line tracking approaches, where the latter are allowed to use the future video frames to predict tracks. 


\subsubsection{Dataset}
The challenge provides seven videos for training along with their ground truth tracks and detected boxes from three detectors. The scenes vary significantly in terms of background, illumination conditions and camera view points. We summarize the main attributes of the provided training data in Table~\ref{tab:detected-boxes}.  
As can be noticed, the provided resolution for scene 05 is different from the others. Similarly, there are also variations in camera frame rates, the average number of objects per frame (i.e. density) and the total number of tracks. Scene 07 is also captured with a lower angle that resulted in significant occlusions. 
All these variations make the dataset  challenging.



\begin{table*}[t]
\centering
\caption{Attributes of MOT17 training data~\cite{MilanL0RS16}. The last three columns indicate the number of detected boxes by the detectors in complete video. `Density' denotes the average number of {\color{\scolor}pedestrians} per frame, and `Move' indicates if the video is recorded by a moving camera (Y) or not (N).} 
\label{tab:detected-boxes}
\begin{tabular}{ccccccccccc}
\hline
\textbf{Video Index}    &\textbf{Resolution}            & \textbf{FPS}    & \textbf{Length (frames)}    & \textbf{Boxes}    & \textbf{Tracks}    &\textbf{Density}    &\textbf{Move}    &\textbf{DPM}~\cite{felzenszwalb2010object}        &\textbf{SDP}~\cite{Yang2016}         &\textbf{FRCNN}~\cite{Ren2017a} \\\hline
02            &$1920\times 1080$    & 30     & 600        & 18581    & 62         &31.0    &N        &7267        &11639        &8186  \\
04            &$1920\times 1080$    & 30     & 1050        & 47557    & 83         &45.3    &N        &39437        &37150        &28406     \\
05            &$640\times 480$    & 14     & 837        & 6917     & 133         &8.3    &Y        &4333        &4767        &3848     \\
09            &$1920\times 1080$    & 30     & 525        & 5325    & 26          &10.1    &N        &5976        &3607        &3049     \\
10            &$1920\times 1080$    & 30     & 654        & 12839    & 57          &19.6    &Y        &8832        &9701        &10371     \\
11            &$1920\times 1080$    & 30     & 900        & 9436    & 75          &15.5    &Y        &8590        &7509        &6007     \\
13            &$1920\times 1080$    & 25     & 750        & 11642    & 110         &8.3    &Y        &5355        &7744        &8442     \\\hline
\end{tabular}
\vspace{-2mm}
\end{table*}

The testing data consists of clips from the same seven scenes, excluding  the training video clips. There are $17,757$ frames in testing data, and it is public knowledge that there are $2,355$ tracks in those clips with $564,228$ boxes. However, the track labels and boxes are not available publicly, and evaluation is performed by the host server in private.


\subsubsection{Evaluation metrics}
\label{secEM}
We benchmark our approach comprehensively using twelve standard evaluation metrics, that include both CLEAR MOT metrics~\cite{Bernardin2008}, and the MT/ML metrics~\cite{Ristani2016}. We summarize these metrics in Table~\ref{tab:mot17-metric}. Definitions of MOTA and MOTAL are provided below:
\begin{align}
\text{MOTA} = 1 - \frac{\sum\limits_t(\text{FP}_t+ \text{FN}_t + \text{ID\_Sw}_t)}{\sum\limits_t\text{GT}_t}.
\label{eq:MOTA}
\end{align}
\begin{align}
\text{MOTAL} = 1 - \frac{\sum\limits_t\left(\text{FP}_t+\text{FN}_t+\log_{10}(\text{ID\_Sw}_t+1)\right)}{\sum\limits_t \text{GT}_t}.
\label{eq:MOTAL}
\end{align}

In the above equations, the subscript `$t$' indicates that the values are computed at the $t^{\text{th}}$ time stamp, whereas `GT' stands for  ground truth. We refer to Table~\ref{tab:mot17-metric} for the other symbols used in the equations.

\begin{table}[t]
\centering
\caption{Metrics used for benchmarking.}
\label{tab:mot17-metric}
\tabcolsep=0.03cm
\begin{tabular}{l c  c p{2.3in}}								\hline
\textbf{Metric} & \textbf{Better} & \textbf{Perfect} &~~~~~~~~~~~~~~~~~~\textbf{Description}                    \\ \hline \hline
MOTA    & higher & 100\%   & Overall Tracking Accuracy. See Eq.~(\ref{eq:MOTA}).                \\ \hline
MOTAL    & higher & 100\%  & Log Tracking Accuracy. See Eq.~(\ref{eq:MOTAL}). \\ \hline
MOTP    & higher & 100\%   & Percentage alignment of predicted bounding box and  ground truth.\\ \hline
Rcll    & higher & 100\%   & The percentage of detected targets. \\ \hline
IDF1    & higher & 100\%   & F1 score of the predicted identities.\\ \hline
MT      & higher & 100\%   & Mostly tracked targets. The percentage of ground-truth trajectories covered by a track hypothesis for 80\% of their life or more.\\ \hline
ML      & lower  & 0     & Mostly lost targets. The percentage of ground-truth trajectories covered by a track hypothesis for 20\% of their life or less.\\ \hline
FP      & lower  & 0       & Number of false positives.                \\ \hline
FN      & lower  & 0       & Number of false negatives.\\ \hline
ID\_Sw.  & lower  & 0       & Identity switches, see \cite{Li2009} for details.                \\ \hline
Frag	& lower  & 0	   & The count of trajectory fragmentations. \\ \hline
Hz      & higher & Inf.    & Processing speed in frames per second.                \\\hline
\end{tabular}
\end{table}


\subsubsection{Results}
\label{sec:Results_MOT17}
We train our DAN with the training data provided by the hosting server. Upon submission, our approach was benchmarked by the server itself. In Table~\ref{tab:benchmark_mot17}, we summarize  current results of the published techniques on the leader board taken directly from the challenge server. Whereas our method is \textit{on-line}, the table also includes  results of {\color{\scolor} the} best performing \textit{off-line}  methods to highlight  competitive performance of our approach.  As can be seen, the proposed approach (named DAN after the network) is able to outperform the existing on-line and off-line methods on five metrics, whereas the performance  generally remains competitive on the other metrics. In particular, our results are significantly better than the existing on-line methods for MOTA and MOTAL that are widely accepted as comprehensive multiple object tracking metrics.

\begin{figure*}[t]
\centering
\includegraphics[width=6in]{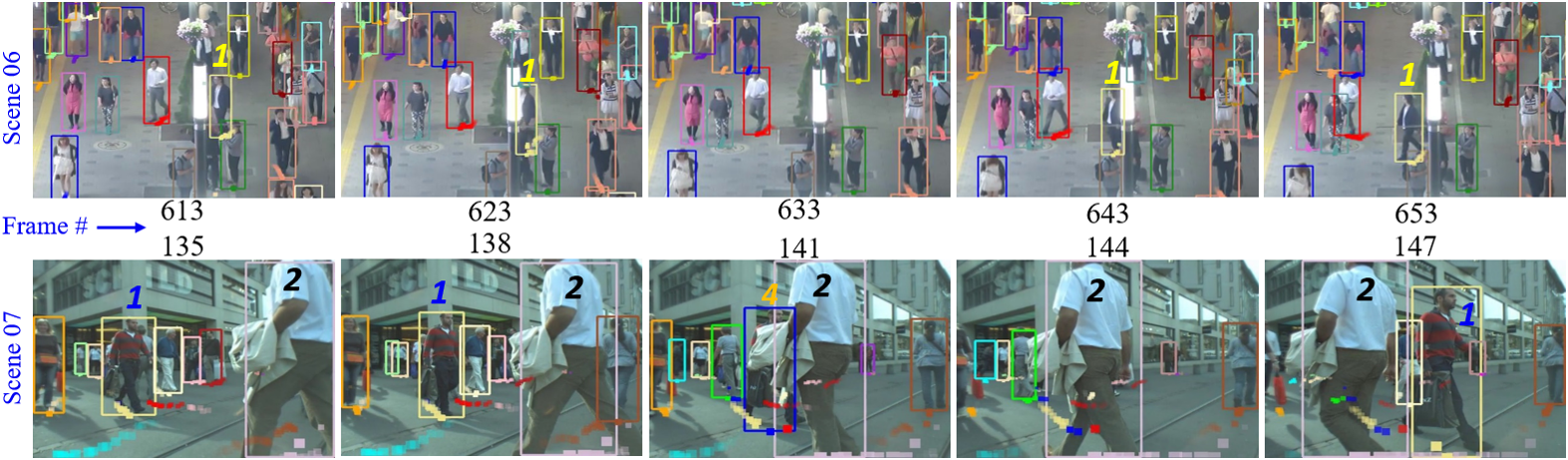}
\caption{Tracking example of the proposed method from MOT17 (taken from the host server). The predicted tracks are identified by the color of bounding boxes. The mentioned identity numbers are for reference in the text only. In both scenes, our approach successfully tracks identity-1 despite inter-frame occlusions. Frame 141 of Scene 07 causes a temporary mis-identification, however, our approach is able to recover well due to deep track association. }
\label{fig:mot17-examples}
\end{figure*}

In Fig.~\ref{fig:mot17-examples}, we show two examples of our method's tracking performance on  MOT17 challenge. The color of bounding boxes in the shown frames indicate the trajectory identity predicted by our tracker. The numbers mentioned on the frames are for reference in the text only. 
The figure presents typical examples of inter-frame occlusions occurring in tracking datasets. In scene 06 (first row), identity-1 disappears in frame 633 (and adjacent frames - not shown), and then reappears in frame 643. Our tracker is able to easily recover from this occlusion (the same color of bounding boxes). Similarly, the occlusion of identity-1 in frame 144 of scene 07 is also handled well by our approach. Our tracker temporarily misjudges the trajectory in frame 141 by assigning it a wrong identity, i.e. 4 due to severe partial occlusion. However, it is able to quickly recover from this situation by looking deeper into the previous frames.  




\begin{table*}[t]
\caption{MOT17 challenge results from the server: The symbol $\uparrow$ indicates that higher values are better, and $\downarrow$ implies lower values are favored. The proposed method is \textit{online}. Offline methods are included for reference only.}
\label{tab:benchmark_mot17}
\tabcolsep=4pt
\centering
\begin{tabular}{lccccccccccccc}
    \hline
    \textbf{Tracker}    &\textbf{Type}		& \textbf{MOTA}$\uparrow$ & \textbf{MOTAL}$\uparrow$ & \textbf{MOTP}$\uparrow$ & \textbf{Rcll}$\uparrow$ & \textbf{IDF1}$\uparrow$ & {\color{\scolor}\textbf{MT}$\uparrow$} & \textbf{ML}$\downarrow$ & \textbf{FP}$\downarrow$ & \textbf{FN}$\downarrow$ & \textbf{ID\_Sw}$\downarrow$ & \textbf{Frag}$\downarrow$ &\textbf{Hz}$\uparrow$    \\ \hline
  FWT\_17~\cite{Henschel2018a}	& offline      	& 51.3173 			& 51.786  			& 77.0024 			& 56.0583 			& 47.5597 			& 21.4 			& 35.3			& 24101 			& 247921 			& 2648			& 4279			& 0.2   \\
jCC~\cite{KeuperTYABS16}		&offline      	& 51.1614 			& 51.4802 			& 75.9164 			& 56.0777 			& \textbf{54.4957} 	& 20.9			& 37.0			& 25937 			& 247822 			& \textbf{1802}	& 2984			& 1.8   \\
MHT\_DAM~\cite{Kim2015}			&offline      	& 50.7132 			& 51.1228 			& \textbf{77.5219} 	& 55.1777 			& 47.1803 			& 20.8			& 36.9			& 22875				& 252889 			& 2314			& \textbf{2865}	& 0.9   \\
EDMT17~\cite{Chen2017}			&offline      	& 50.0464 			& 50.4471 			& 77.2553 			& 56.1689 			& 51.2532 			& \textbf{21.6}			& 36.3			& 32279 			& 247297 			& 2264			& 3260			& 0.6   \\
MHT\_bLSTM~\cite{kim2018multi}	&offline		& 47.5226			& 47.8887			& 77.4943			& 52.494			& 51.9188			& 18.2			& 41.7			& 25981				& 268042			& 2069			& 3124			& 1.9 	\\
IOU17~\cite{Bochinski2017a}		&offline 		& 45.4765			& 46.5371			& 76.8505			& 50.0814			& 39.4037			& 15.7			& 40.5			& \textbf{19993}	& 281643			& 5988			& 7404			& \textbf{1522.9}\\
PHD\_GSDL17~\cite{Fu2017}		&\textit{online}& 48.0439 			& 48.7518 			& 77.1522 			& 52.8641 			& 49.6294 			& 17.1			& 35.6			& 23199 			& 265954 			& 3998 			& 8886			& 6.7   \\
EAMTT~\cite{Sanchez2016}      	&\textit{online}& 42.6344 			& 43.4291 			& 76.0305 			& 48.8728  			& 41.7654 			& 12.7			& 42.7			& 30711 			& 288474 			& 4488			& 5720			& 1.4   \\
GMPHD\_KCF~\cite{Kutschbach2017}&\textit{online}& 39.5737 			& 40.603  			& 74.5414 			& 49.6253 			& 36.6362 			& 8.8			& 43.3			& 50903 			& 284228 			& 5811			& 7414			& 3.3  	\\
GM\_PHD~\cite{Eiselein2012}		&\textit{online}& 36.3560			& 37.1719			& 76.1957			& 41.3771			& 33.9243			& 4.1	& 57.3			& 23723				& 330767			& 4607			& 11317			& 38.4	\\
\textbf{DAN} (Proposed)			&\textit{online}& \textbf{52.4224}	& \textbf{53.916}  	& 76.9071			& \textbf{58.4225} 	& 49.4934 			& 21.4			& \textbf{30.7}	& 25423				& \textbf{234592}	& 8431			&14797			& 6.3 	\\\hline
    \end{tabular}%
\end{table*}


\vspace{-2mm}
\subsection{Multiple Object Tracking 15 (MOT15)}
{\color{\scolor}To} comprehensively benchmark our technique for pedestrian tracking, we also evaluate it on Multiple Object Tracking 15 (MOT15) challenge~\cite{Leal-Taixe2015} {\color{\scolor}that}  deals with multiple pedestrian tracking similar to MOT17. However, due to its earlier release in 2014, MOT15 benchmarks more methods as compared to MOT17.
MOT15 provides 11 video sequences along with their ground truth tracks and detections using the detector proposed by Dollar et al.~\cite{Dollar2014}. The provided ground truth detections do not consider object occlusions and a bounding box for a completely occluded object of interest is provided for training anyway. Such training data can be misleading for our appearance modeling based tracker. Hence, we did not train or fine-tune the DAN with MOT15 training data. Instead, we directly applied our MOT17 model to the MOT15 challenge. The hosting server {\color{\scolor}computed the results of our method using its own  detector~\cite{Dollar2014}}.

\vspace{-2mm}
\subsubsection{Results}
There are 4 video sequences (Venice-1, ADL-Rundle-1, ADL-Rundle-3 and ETH-Crossing) in MOT15 that contain the same scenes as the video sequences 01, 06, 07 and 08 in the MOT17. For a fair comparison of our technique with the benchmarked approaches, we report the average results on these scenes in Table~\ref{tab:benchmark_mot15}. All {\color{\scolor}of the}  approaches in the table contain the training video clips of the same four scenes in their training data. As can be seen, the proposed approach significantly outperforms the existing methods evaluated on the MOT15 challenge on three metrics, especially on the MOTA metric. 

{\color{\scolor} One particularly noticeable method in Table~\ref{tab:benchmark_mot15} is CDA-DDAL~\cite{Baea} that also uses deep features for appearance modeling, but computes appearance affinities using $\ell_2$-distance. Along appearance affinities, it additionally uses shape and motion affinities with the Hungarian algorithm for tracklet association. Despite the use of additional affinities under a related pipeline,  CDA-DDAL significantly underperforms as compared to our tracker. Our technique mainly achieves the performance gain due to effective simultaneous appearance modeling and affinity prediction with DAN. We emphasize that our tracker only uses appearance affinities predicted by DAN. }


\begin{table*}[t]
\caption{MOT15 challenge results: The symbols $\uparrow$ and $\downarrow$ respectively indicate that higher and lower values are preferred. }
\label{tab:benchmark_mot15}
\tabcolsep=4pt
\centering
\begin{tabular}{lcccccccccccc}
    \hline
    \textbf{Tracker}                 & \textbf{Type} & \textbf{MOTA $\uparrow$} & \textbf{MOTP $\uparrow$} & \textbf{IDF1 $\uparrow$} & {\color{\scolor}\textbf{MT $\uparrow$}} & \textbf{ML $\downarrow$} & \textbf{FP $\downarrow$} & \textbf{FN $\downarrow$} & \textbf{ID\_Sw $\downarrow$} & \textbf{Frag $\downarrow$} & \textbf{HZ $\uparrow$} \\ \hline
QuadMOT~\cite{Son2017}           & offline       & 29.30                    & \textbf{75.73}           & 40.88                    & 10.48                    & 34.28                    & 1022.25                  & 3461.00                  & 1670.97                     & 2732.67                    & 6.2                    \\
CNNTCM~\cite{Wang2016}           & offline       & 23.03                    & 73.90                    & 34.90                    & 7.93                     & 41.88                    & 1167.25                  & 3607.50                  & 2792.58                     & 2606.37                    & 1.7                    \\
SiameseCNN~\cite{Leal-Taixe2016} & offline       & 28.78                    & 72.90                    & 39.70                    & 10.83                    & 42.20                    & 633.00                   & 3681.50                  & 846.93                      & 2020.80                    & 52.8                   \\
MHT\_DAM~\cite{Kim2015}           & offline       & 27.58                    & 73.53                    & 44.50                    & \textbf{ 21.30}                    & 34.60                    & 1290.25                  & 3223.75                  & 1453.52                     & 2170.06                    & 0.7                    \\
LP\_SSVM~\cite{Wang2017b}         & offline       & 21.18                    & 73.35                    & 36.15                    & 8.70                     & 37.80                    & 1517.00                  & 3396.00                  & 2002.88                     & 2342.29                    & 41.3                   \\
AMIR15~\cite{Sadeghian2017}      & \textit{online}        & 31.88                    & 73.40                    & 44.13                    & 11.73                    & \textbf{22.53}           & 1105.75                  & 3055.75                  & 3273.27                     & 6360.93                    & 1.9                    \\
HybridDAT~\cite{Chopra2005}      & \textit{online}        & 31.48                    & 75.03                    & 46.90                    & 19.70                    & 27.38                    & 1299.00                  & 2869.00                  & 1373.40                     & 3674.18                    & 4.6                    \\
AM~\cite{Chu2017}                & \textit{online}        & 30.23                    & 72.20                    & \textbf{46.98}           & 12.90                    & 46.75                    & 510.50                   & 3711.00                  & \textbf{755.37}             & 4133.45                    & 0.5                    \\
SCEA~\cite{Yoon2016}             & \textit{online}        & 23.85                    & 73.08                    & 32.65                    & 8.03                     & 45.78                    & 751.50                   & 3847.00                  & 2220.33                     & 3157.50                    & 6.8                    \\
RNN\_LSTM~\cite{Milan2016a}       & \textit{online}        & 13.83                    & 72.28                    & 20.78                    & 6.40            & 37.58                    & 1779.75                  & 3615.50                  & 5287.59                     & 7520.59                    & \textbf{165.2}         \\
{\color{\scolor}CDA-DDAL~\cite{Baea}} 				& {\color{\scolor}\textit{online}}	& {\color{\scolor}32.80}	& {\color{\scolor}70.70}	& {\color{\scolor} 38.70}	& {\color{\scolor}9.70}	& {\color{\scolor}42.20}	& {\color{\scolor}4983.00}	& {\color{\scolor}35690.00}	& {\color{\scolor}\textbf{614.00}}	&{\color{\scolor}1583}	&{\color{\scolor}1.20}\\
\textbf{DAN} (Proposed)          & \textit{online}        & \textbf{38.30}           & 71.10                    & 45.60                    & 17.60                    & 41.20                    & 1290.25                  & \textbf{2700.00}         & 1648.08                     & \textbf{1515.60}           & 6.3               \\
\hline    
\end{tabular}%
\end{table*}

\vspace{-2mm}
\subsection{UA-DETRAC}
\label{sec:DETRAC}
The UA-DETRAC challenge~\cite{Wen2015a,Lyu2017} is based on a large-scale tracking dataset for vehicles. It comprises 100 videos that record around 10 hours of vehicle traffic. The recording is made in 24 different locations, and it includes a wide variety of common vehicle types and traffic conditions. The scenes include urban highways, traffic crossings, and T-junctions etc. Overall, the dataset contains about 140k video frames, 8,250 vehicles, and 1,210k bounding boxes. Similar to the MOT challenges, the UA-DETRAC challenge accepts submissions of tracking approaches and the host server evaluates their performance on a separate test data.

\vspace{-2mm}
\subsubsection{Dataset}
We summarize the {\color{\scolor} main}  attributes of the dataset in Table~\ref{tab:ua-detrac}. The table contains information on both training and testing sets. The available videos have a consistent frame size of $540\times960$, and a  frame rate of 25 fps. All the videos are recorded with static cameras, generally installed at high locations. Although in terms of variations this dataset may appear less challenging than MOT datasets, the larger  size of data and the scenes of crossings and junctions make  tracking in this dataset a difficult task.

\begin{table}[t]
\centering
\caption{Attributes of UA-DETRAC dataset~\cite{Wen2015a,Lyu2017}. The `Boxes' column indicates the total number of bounding boxes in  videos. `Length' is given in  number of frames.}
\label{tab:ua-detrac}
\tabcolsep=0.06cm
\begin{tabular}{ccccccccccc}
\hline
\textbf{Type}        &\textbf{Videos}    & \textbf{Length}   & \textbf{Boxes }       & \textbf{Vehicles}    & \textbf{Tracks}    & \textbf{Density}        \\\hline
Training    &60        & 84k        &578k        &5936        &5.9k        &6.88            \\
Testing        &40        & 56k        &632k        &2314        &2.3k        &11.29            \\\hline
\end{tabular}%
\end{table}%


\subsubsection{Evaluation metrics}
The evaluation metrics used by UA-DETRAC are similar to those introduced in  Table~\ref{tab:mot17-metric}, however, they are computed slightly differently 
{\color{\scolor} using Precision-Recall curve, which is indicated with the prefix `PR' in the table.
As an example}, to compute PR-MOTA, the thresholds of the detectors are gradually varied to compute a 2D precision-recall curve. Then, for each point on the plot, MOTA value is estimated to get a 3D curve. The PR-MOTA is computed as the integral score of the resulting curve. The same procedure is also adopted for computing the other metrics.



\subsubsection{Results}
We {\color{\scolor} report} the quantitative results of our approach in the last row of Table~\ref{tab:result-ua}. The table also summarizes {\color{\scolor} results of other}  top approaches on the leader board for UA-DETRAC challenge at the time of submission of this work. {\color{\scolor}The method names (first column) include the used detectors}, e.g. we used the EB detector~\cite{Wang2017a}, hence EB+DAN. {\color{\scolor}We note that for every technique in the Table (including ours), the choice of detector is empirical. We also tested our tracker with CompACT and RCNN detectors. The values of PR-MOTA and PR-MOTAP metrics for CompACT+DAN are $18.6$ and $35.8$, respectively.  For RCNN+DAN, these values are $15.1$ and $37.1$. The overall performance of our tracker improves with the accuracy of the detector. In the context of detection-based tracking, this implicates an accurate tracking component of the overall technique. These results}  ascertain the overall effectiveness of the proposed approach in tracking vehicles on roads.  

\begin{figure*}[t]
\centering
\includegraphics[width=6in]{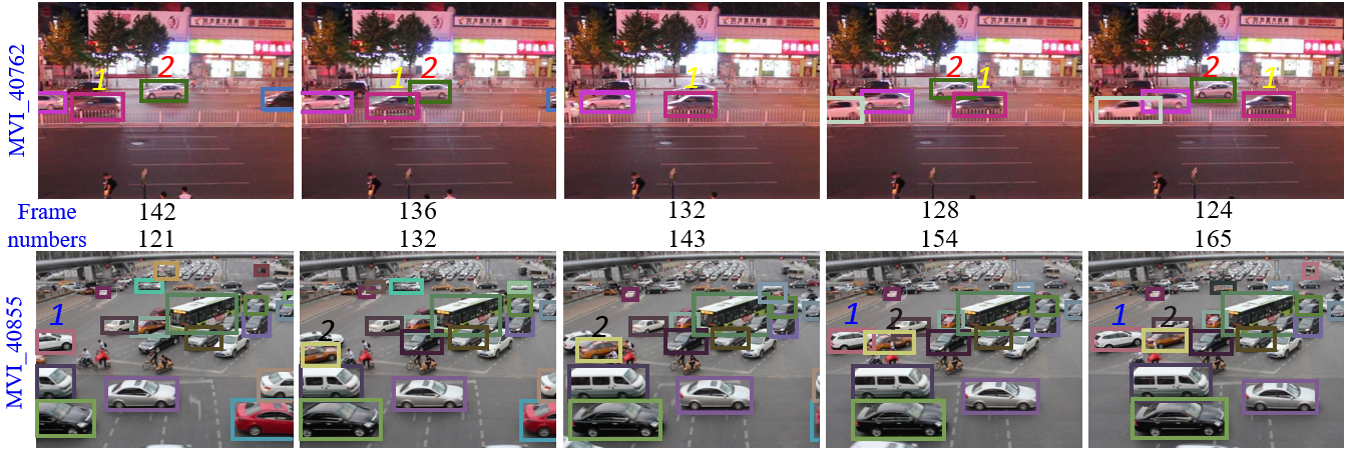}
\caption{Tracking examples from the UA-DETRAC challenge. Predicted tracks are identified by the bounding box colors and the identity numbers are for reference only. The proposed tracker is able to assign identities to their correct track (in both cases) despite missed detection in several frames due to limited performance of the detector for occluded objects.}
\label{fig:ua-examples}
\vspace{-3mm}
\end{figure*}

We also illustrate tracking results of our approach in Fig.~\ref{fig:ua-examples} with the help of two examples. Again, the predicted trajectories are specified by the colors of  bounding boxes, and the mentioned identity numbers are only for referencing in the text. 
In the top row (scene MVI\_40762) the EB detector fails to detect identity-2 in frame 130 - 134 (only frame 132 is shown) due to occlusion by identity-1. However once the detection is made again, our tracker is able to assign the object correctly to its trajectory from frame 136 onward. Similarly, in scene MVI\_40855, the detector is unable to detect identity-1 in frames 132-143 due to occlusion by identity-2. Nevertheless, when the identity-1  is detected again in frame 154, our approach assigns it to its correct trajectory. These examples demonstrate robustness of our approach to missed detections under tracking-by-detection framework.


\begin{table*}[t]
\centering
\caption{UA-DETRAC challenge results: The symbol $\uparrow$ indicates that higher values are better, and $\downarrow$ implies lower values are favored. The names of approaches also include the used detectors. }
\label{tab:result-ua}
\tabcolsep=6pt
\begin{tabular}{lcccccccc}
\hline
\textbf{Name}												&\textbf{PR-MOTA} $\uparrow$ &\textbf{PR-MOTP} $\uparrow$ &\textbf{PR-MT} $\uparrow$  &\textbf{PR-ML} $\downarrow$ &\textbf{PR-FP} $\downarrow$ &\textbf{PR-FN}  $\downarrow$ &\textbf{PR-ID\_Sw} $\downarrow$   &\textbf{Hz} $\uparrow$ \\ \hline
EB~\cite{Wang2017a}+IOUT~\cite{Bochinski2017a}   			&19.4     		&28.9    			&\textbf{17.7}   	&18.4            	&14796.5            &171806.8            	&2311.3       		&\textbf{6902.1}\\ 
R-CNN~\cite{Girshick2014}+IOUT~\cite{Bochinski2017a}		&16.0   		&\textbf{38.3}   	&13.8            	&20.7	            &22535.1            &193041.9            	&5029.4 			&-\\
CompACT~\cite{Cai2015}+GOG~\cite{Pirsiavash2011}			&14.2    		&37.0    			&13.9            	&19.9            	&32092.9            &180183.8            	&3334.6        		&389.5\\
CompACT~\cite{Cai2015}+CMOT~\cite{Bae2014}					&12.6    		&36.1    			&16.1            	&18.6            	&57885.9            &167110.8            	&\textbf{285.3}  	&3.8\\
CompACT~\cite{Cai2015}+H2T~\cite{Wen2014}					&12.4			&35.7				&14.8				&19.4				&51765.7			&173899.8				&852.2				&3.0\\
RCNN~\cite{Girshick2014}+DCT~\cite{Andriyenko2012}			&11.7			&38.0				&10.1				&22.8				&336561.2			&210855.6				&758.7				&0.7\\
CompACT~\cite{Cai2015}+IHTLS~\cite{Dicle2013}				&11.1			&36.8				&13.8				&19.9				&53922.3			&180422.3				&953.6				&19.8\\
CompACT~\cite{Cai2015}+CEM~\cite{Andriyenko2011}			&5.1			&35.2				&3.0				&35.3				&12341.2			&260390.4				&267.9				&4.6\\
Proposed ({EB}~\cite{Wang2017a}+\textbf{DAN})					&\textbf{20.2} &26.3    			&14.5           	&\textbf{18.1}   &\textbf{9747.8}    &\textbf{135978.1}    	&518.2           	&6.3\\\hline

\end{tabular}
\end{table*}



\begin{figure*}[t]
\centering
\includegraphics[width=6in]{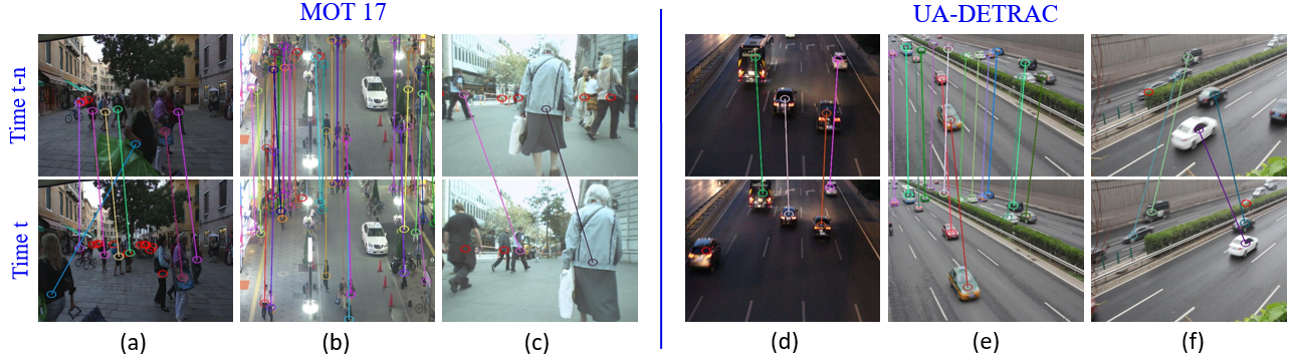}
\caption{Illustration of cross-frame associations based on DAN outputs. The frame pairs are randomly chosen $n\sim[1, 30]$ time-stamps apart. The association remains robust to illumination conditions, partial occlusions and existence of multiple similar looking objects in the video frames. } 
\label{fig:match-examples}
\vspace{-3mm}
\end{figure*}

\vspace{-2mm}
\section{Discussion}
\label{sec:Disc}
{\color{\scolor} {\color{\ncolor} The strength of our approach comes from comprehensive appearance modeling} and effective affinity computation, such that  the latter is fully tailored to the former under end-to-end training of DAN. DAN is the first deep network that models objects' appearance and computes inter-frame object affinities \emph{simultaneously}. This unique property  of our network also sets its performance apart form other methods.}
In Fig.~\ref{fig:match-examples}, we illustrate the object association {\color{\scolor} abilities} of DAN under  practical conditions. Each column of the figure shows a pair of frames that are $n$ time stamps apart, where $n$ is randomly sampled from [1, 30]. The figure shows association between the objects as computed by DAN. Firstly, it can be noticed that the association is robust to large illumination variations. Secondly, DAN is able to comfortably handle significant object occlusions. Additionally, despite the existence of multiple similar looking objects, the network is able to correctly associate the objects. For instance, see Fig.~\ref{fig:match-examples}(e) where there are multiple white and red cars but this does not cause {\color{\scolor}any} problem.
The chosen examples in Fig.~\ref{fig:match-examples} are   random. We observed similar level of performance by DAN for all the cases we tested. 

\begin{figure}[t]
\centering
\includegraphics[width=2in]{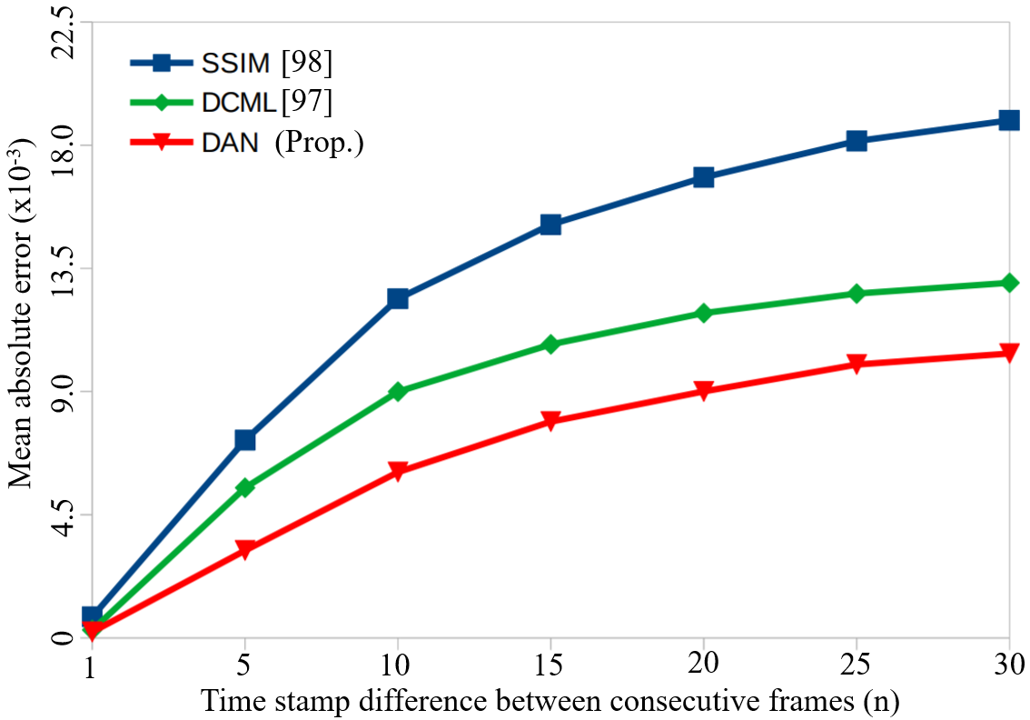}
\caption{{\color{\scolor}Mean absolute error of predicted data association on MOT17. All possible frame pairs are used for each $n$ - the time stamp difference between the two frames in a pair.}} 
\label{fig:MAE}
\vspace{-3mm}
\end{figure}

{\color{\scolor} To evaluate the accuracy of the affinity matrix predicted by DAN, we compare it to more specialized methods for affinity computation, DCML~\cite{wojke2018deep} and SSIM~\cite{wang2004image}. We compute affinities between two frames that are `$n$' time stamps apart in MOT17 dataset, where $n = 1, 5, 10, 15, 20, 25, 30$. The predicted affinity matrices are then used for data association. In Fig.~\ref{fig:MAE}, we plot the (element-wise) mean absolute error between the predicted and the  ground truth association matrices for all possible frame pairs of MOT17 for each value of `$n$'. The consistent low error of DAN  demonstrates the advantage of using a deep network with specialized loss for affinity prediction.}



{\color{\scolor} {\color{\ncolor}The only noticeable situation where DAN underperformed in terms of data association was when the frames contained similar looking objects at very close locations in the scene at multiple time stamps. This sometimes resulted in ID switches between those objects. However, due to the Deep Track Association (Section~\ref{sec:rec}), the proposed method was able to recover well from such scenarios.  
Since our technique is essentially an appearance modeling based tracker, it can be expected to underperform in conditions that are inherently unsuitable for this tracking paradigm, e.g.~highly crowded scenes with multiple similar looking objects. Nevertheless, this issue is associated with all appearance based trackers. }}
 
For an effective architecture of our network, we tested numerous intuitive alternatives. We present a few interesting choices out of those as an ablation analysis here. We choose these cases for their ability to add to our understanding the role of important components in the final network. We introduce DAN-Remove model, that removes the `Feature dimension reduction' layers in Fig.~\ref{fig:sst} and directly concatenates the features to estimate object  affinities. As another case, we  replace the `Compression network' in  affinity estimator component by a single convolutional layer. Hence, the resulting DAN-Replace model maps object features to affinities abruptly instead of doing it gradually.  In another choice, instead of defining our ensemble loss with the \emph{max} operation, we use the \emph{mean} operation, and refer to the resulting network as DAN-Mean.
{\color{\scolor}We also introduce DAN-Curtail that removes the `Extension' sub-network from the `Feature extractor' component of the overall network.}

{\color{\scolor}In Table~\ref{tab:loss}, we provide the MOT17 training loss values at different epochs for the different variants of DAN as well as the MOTA and IDF1 scores at 120 epochs. The MOTA and IDF1 scores are different here from Table \ref{tab:mot17-metric} where the scores on test data from the challenge server are listed. From the table, it} is apparent that the proposed DAN is able to {\color{\scolor} achieve better results} in fewer epochs compared to all the other choices. The closest performance is achieved by DAN-Mean that has the same architecture as DAN but a slightly different loss function. The importance of feature compression  is clear from the poor {\color{\scolor}performance}  of DAN-Remove. Similarly, from the results of DAN-Replace, it is also apparent that gradual compression of feature maps for affinity estimation is more desirable than an abrupt compression.   {\color{\scolor} Finally, the contribution of the Extension sub-network in the overall model is evident from the significant drop in performance of DAN-Curtail.}

\begin{table}[t]
\centering
\caption{{\color{\scolor}Loss values of DAN variants at 50-120 epochs on MOT17 while training. MOTA and IDF1 scores are computed at 120 epochs.}}
\label{tab:loss}
\tabcolsep=4pt
\begin{tabular}{lccccc|c|c}
\hline
\textbf{Variant}     	&\textbf{50}       &\textbf{80}       &\textbf{100}			&\textbf{110}     	&\textbf{120} & {\color\scolor\textbf{MOTA}} & {\color\scolor\textbf{IDF1}}\\\hline
DAN-Replace	    &0.155    &0.115  	&0.108        	&0.112     	&0.111 & {\color\scolor 52.1} & {\color\scolor 48.5}\\
DAN-Remove    	&0.209    &0.181    &0.169        	&0.131     	&0.124 & {\color\scolor 51.7} & {\color\scolor 46.2}\\
DAN-Mean    	&0.134    &0.083    &0.080        	&0.076     	&0.075 & {\color\scolor 53.4} & {\color\scolor 60.7}\\
{\color\scolor DAN-Curtail}   & {\color\scolor 0.367}   & {\color\scolor 0.271}    & {\color\scolor 0.233}        	& {\color\scolor 0.212}     	& {\color\scolor 0.209} & {\color\scolor 45.2} & {\color\scolor 43.7} \\
{\bf DAN}        	    &{\bf 0.107}    & {\bf 0.057}    & {\bf 0.045}        	& {\bf 0.045}     	& {\bf 0.043} & {\color\scolor {\bf 53.5}} & {\color\scolor {\bf 62.3}} \\\hline
\end{tabular}
\vspace{-3mm}
\end{table}

{\color{\scolor} We report the average runtime performance of the major components of our tracking technique in Table~\ref{tab:time}. Each row of the table shows different number of objects to be tracked, i.e.~$N_m$. The timings are for MOT17 validation set used in our experiments, computed with local NVIDIA GeForce GTX Titan GPU. Using the same GPU, the average time taken by the proposed DAN to process a single pair of frames during training is $188.54$ ms, that translates to $1508.33$ ms for a mini-batch of 8 samples. DAN is the only trainable component in our technique, and it is trained in an end-to-end manner. From the table, we can see that a $4$ times increase in $N_m$ only results in a $1.2$ times increase in the overall runtime of our technique.}

\begin{table}[t]
\centering
\caption{{\color{\scolor}Average runtime (ms) for the major components of our technique for MOT17 validation set. $N_m$ denotes the maximum number of allowed objects in a frame.}}
\label{tab:time}
\tabcolsep=4pt
{\color{\scolor}
\begin{tabular}{cccc}\hline
$N_m$ & Feature Extractor & Affinity Estimator & Track Association \\\hline
20    & 2.23               & 10.55              & 0.14              \\
40    & 2.29               & 10.90              & 0.17              \\
60    & 2.46               & 11.52              & 0.18              \\
80    & 2.50               & 12.83              & 0.19              \\\hline
\end{tabular}
}
\vspace{-3mm}
\end{table}

\vspace{-2mm}
\section{Conclusion}
\label{sec:c}
We presented a multiple object tracker that performs on-line  tracking by associating the objects detected in the current frame with those in multiple previous frames. The tracker derives its strength from our proposed Convolutional Neural Network architecture, referred to as Deep Affinity Network (DAN). The proposed DAN models features of pre-detected objects in the video frames at  multiple levels of abstraction, and infers object affinities across different  frames by analyzing exhaustive permutations of the extracted features. The cross-frame objects similarities and object features are recorded by our approach to trace the trajectories of the object. We evaluated our approach on three on-line multiple object tracking challenges MOT17, MOT15 and UA-DETRAC, using twelve different evaluation metrics. The proposed tracker is able to achieve excellent overall performance with the highest Multiple Object Tracking Accuracy on all the challenges. It also achieved an average speed of 6.3 frames per second, while leading the result board as the best performing on-line tracking approach on many of the evaluation metrics.

\vspace{-2mm}
\section{Acknowledgement}\label{sec:a}
This research was supported by ARC grant DP160101458 and DP190102443, the National Natural Science Foundation of China (Grant No. 61572083), the Joint Found of of Ministry of Education of China (Grant No. 6141A02022610), Key projects of key R \& D projects in Shaanxi (Grant No. 2018ZDXM-GY-047), Team cultivation project of Central University (Grant No. 300102248402) and China Scholarship Council. Mubarak Shah acknowledges the support of ODNI, and IARPA, via IARPA R\&D Contract No.~D17PC00345. The views, findings, opinions, and conclusions or recommendations contained herein are those of the authors and should not be interpreted as necessarily representing the official policies or endorsements, either expressed or implied, of the ODNI, IARPA or the U.S.~Government. The U.S.~Government is authorized to reproduce and distribute reprints for Governmental purpose not withstanding any copyright annotation thereon. The GPU used for this work was donated by NVIDIA Corporation.


%





\ifCLASSOPTIONcaptionsoff
  \newpage
\fi



\vspace{-3mm}

\bibliographystyle{IEEEtran}
\bibliography{bare_jrnl_transmag}

\begin{thebibliography}{10}
\providecommand{\url}[1]{#1}
\csname url@samestyle\endcsname
\providecommand{\newblock}{\relax}
\providecommand{\bibinfo}[2]{#2}
\providecommand{\BIBentrySTDinterwordspacing}{\spaceskip=0pt\relax}
\providecommand{\BIBentryALTinterwordstretchfactor}{4}
\providecommand{\BIBentryALTinterwordspacing}{\spaceskip=\fontdimen2\font plus
\BIBentryALTinterwordstretchfactor\fontdimen3\font minus
  \fontdimen4\font\relax}
\providecommand{\BIBforeignlanguage}[2]{{%
\expandafter\ifx\csname l@#1\endcsname\relax
\typeout{** WARNING: IEEEtran.bst: No hyphenation pattern has been}%
\typeout{** loaded for the language `#1'. Using the pattern for}%
\typeout{** the default language instead.}%
\else
\language=\csname l@#1\endcsname
\fi
#2}}
\providecommand{\BIBdecl}{\relax}
\BIBdecl

\bibitem{Luo2017}
W.~Luo, J.~Xing, A.~Milan, X.~Zhang, W.~Liu, X.~Zhao, and T.-K. Kim,
  ``{Multiple Object Tracking: A Literature Review},''
  \emph{arXiv:1409.7618v4}, pp. 1--18, 2017.

\bibitem{Ren2017a}
S.~Ren, K.~He, R.~Girshick, and J.~Sun, ``{Faster R-CNN: Towards Real-Time
  Object Detection with Region Proposal Networks},'' \emph{IEEE TPAMI},
  vol.~39, no.~6, pp. 1137--1149, 2017.

\bibitem{Impiombato2015}
D.~Impiombato, S.~Giarrusso, T.~Mineo, O.~Catalano, C.~Gargano, G.~La~Rosa,
  F.~Russo, G.~Sottile, S.~Billotta, G.~Bonanno, S.~Garozzo, A.~Grillo,
  D.~Marano, and G.~Romeo, ``{You Only Look Once: Unified, Real-Time Object
  Detection Joseph},'' \emph{Nuclear Instruments and Methods in Physics
  Research, Section A: Accelerators, Spectrometers, Detectors and Associated
  Equipment}, vol. 794, pp. 185--192, 2015.

\bibitem{Redmon2016}
J.~Redmon and A.~Farhadi, ``{YOLO9000: Better, Faster, Stronger},'' 2016.

\bibitem{Felzenszwalb2008a}
P.~Felzenszwalb, D.~McAllester, and D.~Ramanan, ``{A discriminatively trained,
  multiscale, deformable part model},'' in \emph{Proc. CVPR}, 2008.

\bibitem{Hu2012}
W.~Hu, X.~Li, W.~Luo, X.~Zhang, S.~Maybank, and Z.~Zhang, ``{Single and
  multiple object tracking using log-euclidean riemannian subspace and
  block-division appearance model},'' \emph{IEEE TPAMI}, vol.~34, no.~12, pp.
  2420--2440, 2012.

\bibitem{Zhang2014}
L.~Zhang and L.~Van Der~Maaten, ``{Preserving structure in model-free
  tracking},'' \emph{IEEE TPAMI}, vol.~36, no.~4, pp. 756--769, 2014.

\bibitem{Xiang2015}
Y.~Xiang, A.~Alahi, and S.~Savarese, ``{Learning to track: Online multi-object
  tracking by decision making},'' in \emph{Proc. IEEE CVPR}, vol. 2015 Inter,
  2015, pp. 4705--4713.

\bibitem{Henschel2018a}
R.~Henschel, L.~Leal-Taix{\'{e}}, D.~Cremers, and B.~Rosenhahn, ``{Fusion of
  Head and Full-Body Detectors for Multi-Object Tracking},'' \emph{Proc.
  CVPRW}, 2018.

\bibitem{Kim2015}
C.~Kim, F.~Li, A.~Ciptadi, and J.~M. Rehg, ``{Multiple hypothesis tracking
  revisited},'' in \emph{Proc. ICCV}, vol. 2015 Inter, 2015, pp. 4696--4704.

\bibitem{Zhang2016b}
S.~Zhang, X.~Lan, H.~Yao, H.~Zhou, D.~Tao, and X.~Li, ``{A Biologically
  Inspired Appearance Model for Robust Visual Tracking},'' \emph{IEEE
  Transactions on Neural Networks and Learning Systems}, 2016.

\bibitem{Nam2015}
H.~Nam and B.~Han, ``{Learning Multi-Domain Convolutional Neural Networks for
  Visual Tracking},'' \emph{Cvpr}, pp. 4293--4302, 2015.

\bibitem{Bae2014}
S.~H. Bae and K.~J. Yoon, ``{Robust online multi-object tracking based on
  tracklet confidence and online discriminative appearance learning},''
  \emph{Proc. CVPR}, pp. 1218--1225, 2014.

\bibitem{Breitenstein2009}
M.~D. Breitenstein, F.~Reichlin, B.~Leibe, E.~Koller-Meier, and L.~Van~Gool,
  ``{Robust tracking-by-detection using a detector confidence particle
  filter},'' \emph{Proc. ICCV}, pp. 1515--1522, 2009.

\bibitem{Fleuret2008}
F.~Fleuret, J.~Berclaz, R.~Lengagne, and P.~Fua, ``{Multicamera people tracking
  with a probabilistic occupancy map},'' \emph{IEEE TPAMI}, vol.~30, no.~2, pp.
  267--282, 2008.

\bibitem{Fu2017}
Z.~Fu, P.~Feng, S.~M. Naqvi, and J.~A. Chambers, ``{Particle PHD filter based
  multi-target tracking using discriminative group-structured dictionary
  learning},'' in \emph{ICASSP}, 2017, pp. 4376--4380.

\bibitem{Kutschbach2017}
T.~Kutschbach, E.~Bochinski, V.~Eiselein, and T.~Sikora, ``{Sequential sensor
  fusion combining probability hypothesis density and kernelized correlation
  filters for multi-object tracking in video data},'' in \emph{Proc. AVSS},
  2017.

\bibitem{Baea}
S.~H. Bae and K.~J. Yoon, ``{Confidence-Based Data Association and
  Discriminative Deep Appearance Learning for Robust Online Multi-Object
  Tracking},'' \emph{IEEE Transactions on Pattern Analysis and Machine
  Intelligence}, vol.~40, no.~3, pp. 595--610, 2018.

\bibitem{Wen2014}
L.~Wen, W.~Li, J.~Yan, Z.~Lei, D.~Yi, and S.~Z. Li, ``{Multiple target tracking
  based on undirected hierarchical relation hypergraph},'' \emph{Proc. CVPR},
  vol.~1, pp. 1282--1289, 2014.

\bibitem{Kuo2010}
C.~H. Kuo, C.~Huang, and R.~Nevatia, ``{Multi-target tracking by on-line
  learned discriminative appearance models},'' in \emph{Proc. CVPR}, 2010, pp.
  685--692.

\bibitem{Izadinia2012}
H.~Izadinia, I.~Saleemi, W.~Li, and M.~Shah, ``{(MP)2T: Multiple people
  multiple parts tracker},'' in \emph{Lecture Notes in Computer Science}, vol.
  7577 LNCS, no. PART 6, 2012, pp. 100--114.

\bibitem{Yamaguchi2011}
K.~Yamaguchi, A.~C. Berg, L.~E. Ortiz, and T.~L. Berg, ``{Who are you with and
  where are you going?}'' in \emph{Cvpr 2011}, 2011, pp. 1345--1352.

\bibitem{Shafique2008}
K.~Shafique, W.~L. Mun, and N.~Haering, ``{A rank constrained continuous
  formulation of multi-frame multi-target tracking problem},'' in \emph{Proc.
  CVPR}, 2008.

\bibitem{Yu2007}
Q.~Yu, G.~Medioni, and I.~Cohen, ``{Multiple Target Tracking Using
  Spatio-Temporal Markov Chain Monte Carlo Data Association},'' in \emph{Proc.
  CVPR}, 2007, pp. 1--8.

\bibitem{yang2012multi}
R.~Nevatia, ``{Multi-target tracking by online learning of non-linear motion
  patterns and robust appearance models},'' in \emph{Proc. CVPR}.\hskip 1em
  plus 0.5em minus 0.4em\relax IEEE, 2012, pp. 1918--1925.

\bibitem{Ning2016}
G.~Ning, Z.~Zhang, C.~Huang, Z.~He, X.~Ren, and H.~Wang, ``{Spatially
  Supervised Recurrent Convolutional Neural Networks for Visual Object
  Tracking},'' 2016.

\bibitem{Munkres1957}
J.~Munkres, ``{Algorithms for the Assignment and Transportation Problems},''
  \emph{Journal of the Society for Industrial and Applied Mathematics}, vol.~5,
  no.~1, pp. 32--38, 1957.

\bibitem{Leal-Taixe2015}
L.~Leal-Taix{\'{e}}, A.~Milan, I.~Reid, S.~Roth, and K.~Schindler,
  ``{MOTChallenge 2015: Towards a Benchmark for Multi-Target Tracking},'' pp.
  1--15, 2015.

\bibitem{MilanL0RS16}
A.~Milan, L.~Leal-Taix{\'{e}}, I.~D. Reid, S.~Roth, and K.~Schindler, ``{MOT16:
  A Benchmark for Multi-Object Tracking},'' \emph{CoRR}, vol. abs/1603.0, 2016.

\bibitem{Wen2015a}
L.~Wen, D.~Du, Z.~Cai, Z.~Lei, M.-C. Chang, H.~Qi, J.~Lim, M.-H. Yang, and
  S.~Lyu, ``{UA-DETRAC: A New Benchmark and Protocol for Multi-Object Detection
  and Tracking},'' 2015.

\bibitem{Lyu2017}
S.~Lyu, M.~C. Chang, D.~Du, L.~Wen, H.~Qi, Y.~Li, Y.~Wei, L.~Ke, T.~Hu,
  M.~Del~Coco, P.~Carcagni, D.~Anisimov, E.~Bochinski, F.~Galasso, F.~Bunyak,
  G.~Han, H.~Ye, H.~Wang, K.~Palaniappan, K.~Ozcan, L.~Wang, L.~Wang, M.~Lauer,
  N.~Watcharapinchai, N.~Song, N.~M. Al-Shakarji, S.~Wang, S.~Amin,
  S.~Rujikietgumjorn, T.~Khanova, T.~Sikora, T.~Kutschbach, V.~Eiselein,
  W.~Tian, X.~Xue, X.~Yu, Y.~Lu, Y.~Zheng, Y.~Huang, and Y.~Zhang, ``{UA-DETRAC
  2017: Report of AVSS2017 {\&} IWT4S Challenge on Advanced Traffic
  Monitoring},'' in \emph{Proc. IEEE AVSS}, 2017.

\bibitem{Emami2018}
P.~Emami, P.~M. Pardalos, L.~Elefteriadou, and S.~Ranka, ``{Machine Learning
  Methods for Solving Assignment Problems in Multi-Target Tracking},'' vol.~1,
  no.~1, pp. 1--35, 2018.

\bibitem{Tian2018}
Y.~Tian, A.~Dehghan, and M.~Shah, ``On detection, data association and
  segmentation for multi-target tracking,'' \emph{IEEE Transactions on Pattern
  Analysis and Machine Intelligence}, pp. 1--1, 2018.

\bibitem{wen2018learning}
L.~Wen, D.~Du, S.~Li, X.~Bian, and S.~Lyu, ``Learning non-uniform hypergraph
  for multi-object tracking,'' \emph{arXiv preprint arXiv:1812.03621}, 2018.

\bibitem{sheng2018heterogeneous}
H.~Sheng, Y.~Zhang, J.~Chen, Z.~Xiong, and J.~Zhang, ``Heterogeneous
  association graph fusion for target association in multiple object
  tracking,'' \emph{IEEE Transactions on Circuits and Systems for Video
  Technology}, 2018.

\bibitem{shafique2005noniterative}
K.~Shafique and M.~Shah, ``A noniterative greedy algorithm for multiframe point
  correspondence,'' \emph{IEEE transactions on pattern analysis and machine
  intelligence}, vol.~27, no.~1, pp. 51--65, 2005.

\bibitem{reid1979algorithm}
D.~Reid \emph{et~al.}, ``An algorithm for tracking multiple targets,''
  \emph{IEEE transactions on Automatic Control}, vol.~24, no.~6, pp. 843--854,
  1979.

\bibitem{shu2012part}
G.~Shu, A.~Dehghan, O.~Oreifej, E.~Hand, and M.~Shah, ``Part-based
  multiple-person tracking with partial occlusion handling,'' in \emph{Proc.
  CVPR}.\hskip 1em plus 0.5em minus 0.4em\relax IEEE, 2012, pp. 1815--1821.

\bibitem{RoshanZamir2012}
A.~Roshan~Zamir, A.~Dehghan, and M.~Shah, ``{GMCP-tracker: Global multi-object
  tracking using generalized minimum clique graphs},'' in \emph{Lecture Notes
  in Computer Science}, 2012.

\bibitem{wu2007detection}
B.~Wu and R.~Nevatia, ``Detection and tracking of multiple, partially occluded
  humans by bayesian combination of edgelet based part detectors,''
  \emph{International Journal of Computer Vision}, vol.~75, no.~2, pp.
  247--266, 2007.

\bibitem{dehghan2015gmmcp}
A.~Dehghan, S.~Modiri~Assari, and M.~Shah, ``Gmmcp tracker: Globally optimal
  generalized maximum multi clique problem for multiple object tracking,'' in
  \emph{Proc. CVPR}, 2015, pp. 4091--4099.

\bibitem{Pirsiavash2011}
H.~Pirsiavash, D.~Ramanan, and C.~C. Fowlkes, ``{Globally-optimal greedy
  algorithms for tracking a variable number of objects},'' \emph{Proc. CVPR},
  pp. 1201--1208, 2011.

\bibitem{butt2013multi}
A.~A. Butt and R.~T. Collins, ``Multi-target tracking by lagrangian relaxation
  to min-cost network flow,'' in \emph{Proc. CVPR}, 2013, pp. 1846--1853.

\bibitem{Berclaz2011}
J.~Berclaz, F.~Fleuret, E.~T{\"{u}}retken, and P.~Fua, ``{Multiple object
  tracking using k-shortest paths optimization},'' \emph{IEEE TPAMI}, vol.~33,
  no.~9, pp. 1806--1819, 2011.

\bibitem{shitrit2014multi}
H.~B. Shitrit, J.~Berclaz, F.~Fleuret, and P.~Fua, ``Multi-commodity network
  flow for tracking multiple people,'' \emph{IEEE TPAMI}, vol.~36, no.~8, pp.
  1614--1627, 2014.

\bibitem{Chari2015}
V.~Chari, S.~Lacoste-Julien, I.~Laptev, and J.~Sivic, ``{On pairwise costs for
  network flow multi-object tracking},'' \emph{Proc. CVPR}, vol. 07-12-June,
  pp. 5537--5545, 2015.

\bibitem{felzenszwalb2010object}
P.~F. Felzenszwalb, R.~B. Girshick, D.~McAllester, and D.~Ramanan, ``{Object
  detection with discriminatively trained part-based models},'' \emph{IEEE
  TPAMI}, vol.~32, no.~9, pp. 1627--1645, 2010.

\bibitem{leibe2008coupled}
B.~Leibe, K.~Schindler, N.~Cornelis, and L.~Van~Gool, ``Coupled object
  detection and tracking from static cameras and moving vehicles,'' \emph{IEEE
  TPAMI}, vol.~30, no.~10, pp. 1683--1698, 2008.

\bibitem{tang2015subgraph}
S.~Tang, B.~Andres, M.~Andriluka, and B.~Schiele, ``Subgraph decomposition for
  multi-target tracking,'' in \emph{Proc. CVPR}, 2015, pp. 5033--5041.

\bibitem{Kalal2011}
Z.~Kalal, K.~Mikolajczyk, and J.~Matas, ``{Tracking-Learning-Detection.}''
  \emph{IEEE TPAMI}, vol.~34, no.~7, pp. 1409--1422, 2011.

\bibitem{hare2011struck}
S.~Hare, A.~Saffari, and P.~H. Torr, ``Struck: Structured output tracking with
  kernels,'' in \emph{Proc. ICCV}.\hskip 1em plus 0.5em minus 0.4em\relax IEEE,
  2011, pp. 263--270.

\bibitem{wang2011superpixel}
S.~Wang, H.~Lu, F.~Yang, and M.-H. Yang, ``Superpixel tracking,'' 2011.

\bibitem{zhang2013structure}
L.~Zhang and L.~van~der Maaten, ``Structure preserving object tracking,'' in
  \emph{Proc. CVPR}, 2013, pp. 1838--1845.

\bibitem{Milan2014}
A.~Milan, S.~Roth, and K.~Schindler, ``{Continuous energy minimization for
  multitarget tracking},'' \emph{IEEE Transactions on Pattern Analysis and
  Machine Intelligence}, vol.~36, no.~1, pp. 58--72, 2014.

\bibitem{Bochinski2017a}
E.~Bochinski, V.~Eiselein, and T.~Sikora, ``{High-Speed tracking-by-detection
  without using image information},'' in \emph{Proc. IEEE AVSS}, 2017.

\bibitem{Chen2017}
J.~Chen, H.~Sheng, Y.~Zhang, and Z.~Xiong, ``{Enhancing Detection Model for
  Multiple Hypothesis Tracking},'' in \emph{2017 IEEE Conference on Computer
  Vision and Pattern Recognition Workshops (CVPRW)}, 2017, pp. 2143--2152.

\bibitem{Bertinetto2016}
L.~Bertinetto, J.~Valmadre, J.~F. Henriques, A.~Vedaldi, and P.~H. Torr,
  ``{Fully-convolutional siamese networks for object tracking},'' \emph{LNCS},
  vol. 9914, pp. 850--865, 2016.

\bibitem{Son2017}
J.~Son, M.~Baek, M.~Cho, and B.~Han, ``{Multi-Object Tracking with Quadruplet
  Convolutional Neural Networks},'' \emph{Cvpr}, pp. 5620--5629, 2017.

\bibitem{Schulter2017}
S.~Schulter, P.~Vernaza, W.~Choi, and M.~Chandraker, ``{Deep Network Flow for
  Multi-Object Tracking},'' \emph{Cvpr}, pp. 6951--6960, 2017.

\bibitem{Feichtenhofer2017}
C.~Feichtenhofer, A.~Pinz, and A.~Zisserman, ``{Detect to Track and Track to
  Detect},'' \emph{Proc. ICCV}, vol. 2017-Octob, pp. 3057--3065, 2017.

\bibitem{Insafutdinov2017}
E.~Insafutdinov, M.~Andriluka, L.~Pishchulin, S.~Tang, E.~Levinkov, B.~Andres,
  and B.~Schiele, ``{ArtTrack: Articulated multi-person tracking in the
  wild},'' \emph{Proc. CVPR}, vol. 2017, pp. 1293--1301, 2017.

\bibitem{Chopra2005}
S.~Chopra, R.~Hadsell, and Y.~LeCun, ``{Learning a similarity metric
  discriminatively, with application to face verification},'' \emph{Proc.
  CVPR}, vol.~1, pp. 539--546, 2005.

\bibitem{Dai2016}
S.~Hershey, S.~Chaudhuri, D.~P. Ellis, J.~F. Gemmeke, A.~Jansen, R.~C. Moore,
  M.~Plakal, D.~Platt, R.~A. Saurous, B.~Seybold, M.~Slaney, R.~J. Weiss, and
  K.~Wilson, ``{CNN architectures for large-scale audio classification},''
  \emph{Proc. ICASSP}, no. Nips, pp. 131--135, 2017.

\bibitem{li2018high}
B.~Li, J.~Yan, W.~Wu, Z.~Zhu, and X.~Hu, ``High performance visual tracking
  with siamese region proposal network,'' in \emph{Proc. CVPR}, 2018, pp.
  8971--8980.

\bibitem{zhang2018learning}
Y.~Zhang, D.~Wang, L.~Wang, J.~Qi, and H.~Lu, ``Learning regression and
  verification networks for long-term visual tracking,'' \emph{arXiv preprint
  arXiv:1809.04320}, 2018.

\bibitem{Milan2016}
A.~Milan, L.~Leal-Taixe, I.~Reid, S.~Roth, and K.~Schindler, ``{MOT16: A
  Benchmark for Multi-Object Tracking},'' pp. 1--12, 2016.

\bibitem{Geiger2012}
A.~Geiger, P.~Lenz, and R.~Urtasun, ``{Are we ready for autonomous driving? the
  KITTI vision benchmark suite},'' in \emph{Proc. CVPR}, 2012, pp. 3354--3361.

\bibitem{Patino2016}
L.~Patino, T.~Cane, A.~Vallee, and J.~Ferryman, ``{PETS 2016: Dataset and
  Challenge},'' in \emph{Proc. CVPRW}, 2016, pp. 1240--1247.

\bibitem{Yang2016}
F.~Yang, W.~Choi, and Y.~Lin, ``{Exploit All the Layers: Fast and Accurate CNN
  Object Detector with Scale Dependent Pooling and Cascaded Rejection
  Classifiers},'' in \emph{Proc. CVPR}, 2016, pp. 2129--2137.

\bibitem{Dollar2014}
P.~Dollar, R.~Appel, S.~Belongie, and P.~Perona, ``{Fast feature pyramids for
  object detection},'' \emph{IEEE Transactions on Pattern Analysis and Machine
  Intelligence}, 2014.

\bibitem{Wang2017a}
L.~Wang, Y.~Lu, H.~Wang, Y.~Zheng, H.~Ye, and X.~Xue, ``{Evolving boxes for
  fast vehicle detection},'' \emph{Proceedings - IEEE International Conference
  on Multimedia and Expo}, pp. 1135--1140, 2017.

\bibitem{Liu2016}
W.~Liu, D.~Anguelov, D.~Erhan, C.~Szegedy, S.~Reed, C.~Y. Fu, and A.~C. Berg,
  ``{SSD: Single shot multibox detector},'' \emph{Lecture Notes in Computer
  Science}, vol. 9905 LNCS, pp. 21--37, 2016.

\bibitem{Simonyan2015}
K.~Simonyan and A.~Zisserman, ``{Very Deep Convolutional Networks for
  Large-Scale Image Recognition},'' \emph{International Conference on Learning
  Representations (ICRL)}, pp. 1--14, 2015.

\bibitem{ioffe2015batch}
S.~Ioffe and C.~Szegedy, ``{Batch normalization: Accelerating deep network
  training by reducing internal covariate shift},'' \emph{arXiv preprint
  arXiv:1502.03167}, 2015.

\bibitem{Nair2010}
V.~Nair and G.~E. Hinton, ``{Rectified Linear Units Improve Restricted
  Boltzmann Machines},'' \emph{Proceedings of the 27th International Conference
  on Machine Learning}, no.~3, pp. 807--814, 2010.

\bibitem{Paszke2017}
A.~Paszke, G.~Chanan, Z.~Lin, S.~Gross, E.~Yang, L.~Antiga, and Z.~Devito,
  ``{Automatic differentiation in PyTorch},'' \emph{Advances in Neural
  Information Processing Systems 30}, no. Nips, pp. 1--4, 2017.

\bibitem{zhang2015deep}
S.~Zhang, A.~E. Choromanska, and Y.~LeCun, ``{Deep learning with elastic
  averaging SGD},'' in \emph{Advances in Neural Information Processing
  Systems}, 2015, pp. 685--693.

\bibitem{Bernardin2008}
K.~Bernardin and R.~Stiefelhagen, ``{Evaluating multiple object tracking
  performance: The CLEAR MOT metrics},'' \emph{Eurasip Journal on Image and
  Video Processing}, vol. 2008, 2008.

\bibitem{Ristani2016}
E.~Ristani, F.~Solera, R.~Zou, R.~Cucchiara, and C.~Tomasi, ``{Performance
  measures and a data set for multi-target, multi-camera tracking},'' in
  \emph{Lecture Notes in Computer Science}, vol. 9914 LNCS, 2016, pp. 17--35.

\bibitem{Li2009}
Y.~Li, C.~Huang, and R.~Nevatia, ``{Learning to associate: Hybridboosted
  multi-target tracker for crowded scene},'' in \emph{Proc. CVPRW}, 2009, pp.
  2953--2960.

\bibitem{KeuperTYABS16}
M.~Keuper, S.~Tang, Y.~Zhongjie, B.~Andres, T.~Brox, and B.~Schiele, ``{A
  Multi-cut Formulation for Joint Segmentation and Tracking of Multiple
  Objects},'' \emph{CoRR}, vol. abs/1607.0, 2016.

\bibitem{kim2018multi}
C.~Kim, F.~Li, and J.~Rehg, ``{Multi-object Tracking with Neural Gating Using
  Bilinear LSTM},'' in \emph{Proc. ECCV}, 2018, pp. 200--215.

\bibitem{Sanchez2016}
R.~Sanchez-Matilla, F.~Poiesi, and A.~Cavallaro, ``{Online Multi-target
  Tracking with Strong and Weak Detections},'' in \emph{Computer Vision -- ECCV
  2016 Workshops}, G.~Hua and H.~J{\'{e}}gou, Eds.\hskip 1em plus 0.5em minus
  0.4em\relax Cham: Springer International Publishing, 2016, pp. 84--99.

\bibitem{Eiselein2012}
V.~Eiselein, D.~Arp, M.~P{\"{a}}tzold, and T.~Sikora, ``{Real-time multi-human
  tracking using a probability hypothesis density filter and multiple
  detectors},'' in \emph{Proc. IEEE AVSS 2012}, 2012, pp. 325--330.

\bibitem{Wang2016}
B.~Wang, L.~Wang, B.~Shuai, Z.~Zuo, T.~Liu, K.~L. Chan, and G.~Wang, ``{Joint
  Learning of Convolutional Neural Networks and Temporally Constrained Metrics
  for Tracklet Association},'' in \emph{Proc. CVPRW}, 2016.

\bibitem{Leal-Taixe2016}
L.~Leal-Taixe, C.~Canton-Ferrer, and K.~Schindler, ``{Learning by Tracking:
  Siamese CNN for Robust Target Association},'' in \emph{Proc. CVPRW}, 2016.

\bibitem{Wang2017b}
S.~Wang and C.~C. Fowlkes, ``{Learning Optimal Parameters for Multi-target
  Tracking with Contextual Interactions},'' \emph{International Journal of
  Computer Vision}, 2017.

\bibitem{Sadeghian2017}
A.~Sadeghian, A.~Alahi, and S.~Savarese, ``{Tracking the Untrackable: Learning
  to Track Multiple Cues with Long-Term Dependencies},'' in \emph{Proc. ICCV},
  2017.

\bibitem{Chu2017}
Q.~Chu, W.~Ouyang, H.~Li, X.~Wang, B.~Liu, and N.~Yu, ``{Online Multi-object
  Tracking Using CNN-Based Single Object Tracker with Spatial-Temporal
  Attention Mechanism},'' in \emph{Proc. ICCV}, 2017.

\bibitem{Yoon2016}
J.~H. Yoon, C.-R. Lee, M.-H. Yang, and K.-J. Yoon, ``{Online Multi-object
  Tracking via Structural Constraint Event Aggregation},'' in \emph{Proc.
  CVPR}, 2016.

\bibitem{Milan2016a}
A.~Milan, S.~H. Rezatofighi, A.~Dick, K.~Schindler, and I.~Reid, ``{Online
  Multi-target Tracking using Recurrent Neural Networks},'' \emph{Arxiv}, 2016.

\bibitem{Girshick2014}
R.~Girshick, J.~Donahue, T.~Darrell, and J.~Malik, ``{Rich feature hierarchies
  for accurate object detection and semantic segmentation},'' in \emph{Proc.
  CVPR}, 2014.

\bibitem{Cai2015}
Z.~Cai, M.~Saberian, and N.~Vasconcelos, ``{Learning complexity-aware cascades
  for deep pedestrian detection},'' in \emph{Proc. ICCV}, vol. 2015 Inter,
  2015, pp. 3361--3369.

\bibitem{Andriyenko2012}
A.~Andriyenko, K.~Schindler, and S.~Roth, ``{Discrete-continuous optimization
  for multi-target tracking},'' in \emph{Proc. CVPR}, 2012, pp. 1926--1933.

\bibitem{Dicle2013}
C.~Dicle, O.~I. Camps, and M.~Sznaier, ``{The way they move: Tracking multiple
  targets with similar appearance},'' in \emph{Proc. ICCV}, 2013, pp.
  2304--2311.

\bibitem{Andriyenko2011}
A.~Andriyenko and K.~Schindler, ``{Multi-target tracking by continuous energy
  minimization},'' in \emph{Proc. CVPR}, 2011, pp. 1265--1272.

\bibitem{wojke2018deep}
N.~Wojke and A.~Bewley, ``Deep cosine metric learning for person
  re-identification,'' in \emph{Proc. WACV}.\hskip 1em plus 0.5em minus
  0.4em\relax IEEE, 2018, pp. 748--756.

\bibitem{wang2004image}
Z.~Wang, A.~C. Bovik, H.~R. Sheikh, E.~P. Simoncelli \emph{et~al.}, ``Image
  quality assessment: from error visibility to structural similarity,''
  \emph{IEEE TIP}, vol.~13, no.~4, pp. 600--612, 2004.

\end{thebibliography}

\begin{IEEEbiography}[{\includegraphics[width=1in,height=1.25in,clip,keepaspectratio]{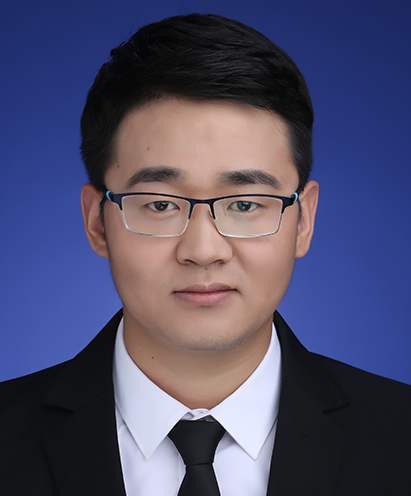}}]{ShiJie Sun} received his B.S. in software engineering from The University of Chang'an University and is currently working towards the Ph.D. degree in intelligent transportation and information system engineering with Chang'an University. Currently, he is a visiting (joint) Ph.D. candidate at the University of Western Australia since October 2017. His research interests include machine learning, object detection, localization \& tracking, action recognition. 
\end{IEEEbiography}

\begin{IEEEbiography}[{\includegraphics[width=1in,height=1.25in,clip,keepaspectratio]{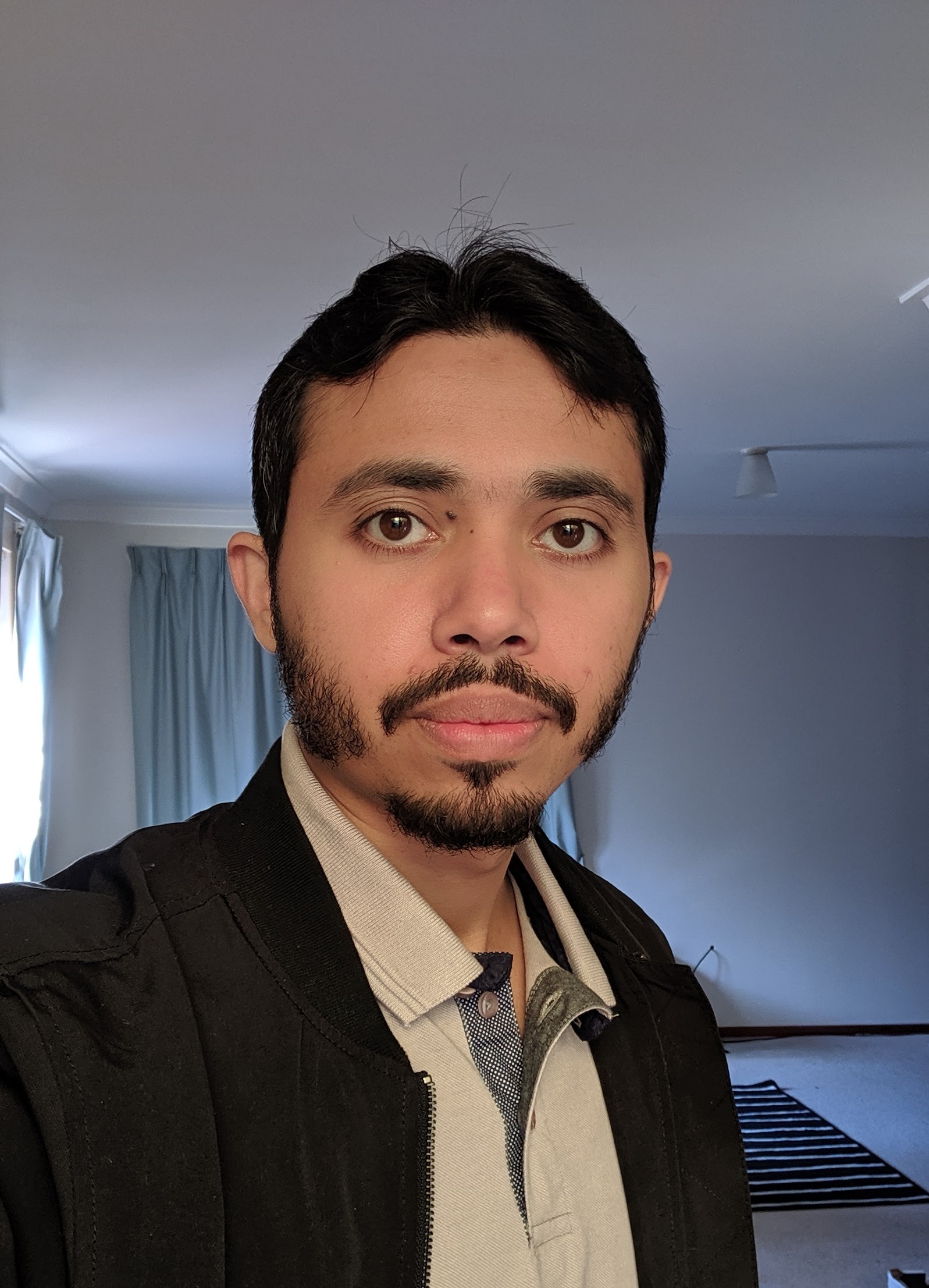}}]{Naveed Akhtar} 
received his PhD in Computer Vision from The University of Western Australia (UWA) and Master degree in Computer Science from Hochschule Bonn-Rhein-Sieg, Germany. His research in Computer Vision is regularly published in   reputed venues of the filed. He also serves as an Associate Editor of IEEE Access.  
Currently, he is a Lecturer at UWA. Previously, he also served as a Research Fellow at UWA and the Australian National University. His  research interests include multiple object tracking, adversarial deep learning, action recognition,  and hyperspectral image analysis.
\end{IEEEbiography}

\begin{IEEEbiography}[{\includegraphics[width=1in,height=1.25in,clip,keepaspectratio]{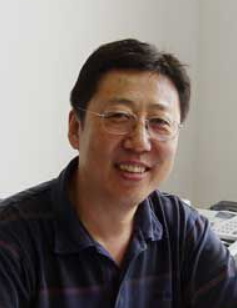}}]{HuanSheng Song} 
received the B.S. and M.S. degrees in communication and electronic systems and the Ph.D. degree in information and communication engineering from Xi’an Jiaotong University, Xi’an, China, in 1985, 1988, and 1996, respectively. Since 2004, he has been with the Information Engineering Institute, Chang’an University, Xi’an, where he became a Professor in 2006 and was nominated as the Dean in 2012. He has been involved in research on intelligent transportation systems for many years and has led a research team to develop a vehicle license plate reader and a traffic event detection system based on videos, which has brought about complete industrialization. His current research interests include image processing, recognition  and tracking as well as intelligent transportation systems.
\end{IEEEbiography}

\begin{IEEEbiography}[{\includegraphics[width=1in,height=1.25in,clip,keepaspectratio]{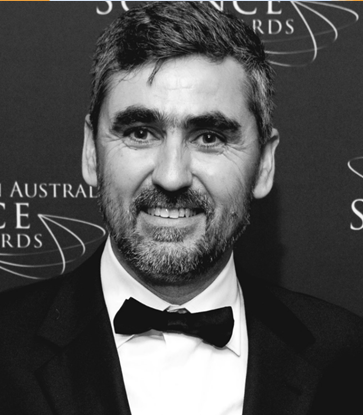}}]{Ajmal Mian} is a Professor of Computer Science at The University of Western Australia. He has received two prestigious fellowships and several research grants from the Australian Research Council and the National Health and Medical Research Council of Australia with a combined funding of over \$12 million. He was the West Australian Early Career Scientist of the Year 2012 and has received several awards including the Excellence in Research Supervision Award, EH Thompson Award, ASPIRE Professional Development Award, Vice-chancellors Mid-career Award, Outstanding Young Investigator Award, the Australasian Distinguished Dissertation Award and various best paper awards. His research interests are in computer vision, machine learning, 3D shape analysis, face recognition, human action recognition and video analysis.
\end{IEEEbiography}

\begin{IEEEbiography}[{\includegraphics[width=1in,height=1.25in,clip,keepaspectratio]{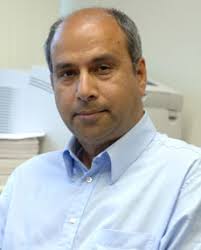}}]{Mubarak Shah} the trustee chair professor of computer science, is the founding director of the Center for Research in Computer Vision at University of Central Florida. He is an editor of
an international book series on video computing,
was editor-in-chief of Machine Vision and
Applications journal, and an associate editor of
ACM Computing Surveys journal. His research
interests include video surveillance, visual tracking,
human activity recognition, visual analysis of
crowded scenes, video registration, UAV video
analysis, and so on. He is a fellow of the IEEE, AAAS, IAPR, and SPIE.
\end{IEEEbiography}

\end{document}